\documentclass[a4paper,fleqn]{cas-dc}

\let\thead\relax

\usepackage[T1]{fontenc}
\usepackage{hyperref}
\usepackage{url}
\usepackage{booktabs}
\usepackage{colortbl}
\usepackage{amsfonts}
\usepackage{nicefrac}
\usepackage{mathtools,bm,amssymb,xcolor}
\definecolor{MetricBlue}{HTML}{78B7C9}
\usepackage{subcaption,placeins}
\usepackage{adjustbox,multirow,dcolumn}
\usepackage{algorithm,algpseudocode,setspace}
\usepackage{xstring}
\usepackage{cleveref}
\usepackage{graphicx}
\usepackage[authoryear]{natbib}
\setcitestyle{authoryear,round}
\usepackage{comment}
\usepackage{url}

\usepackage{amsmath,amsfonts,bm}

\def\1{\bm{1}}

\DeclareMathAlphabet{\mathsfit}{\encodingdefault}{\sfdefault}{m}{sl}
\SetMathAlphabet{\mathsfit}{bold}{\encodingdefault}{\sfdefault}{bx}{n}

\makeatletter
\DeclareRobustCommand\onedot{\futurelet\@let@token\@onedot}
\def\@onedot{\ifx\@let@token.\else.\null\fi}

\graphicspath{{figures/}}

\newcolumntype{d}{D{.}{.}{3.2}}
\makeatletter
\newcolumntype{B}{>{\boldmath\DC@{.}{.}{3.2}}c<{\DC@end}}
\makeatother

\newcommand{\thead}[1]{\multicolumn{1}{c}{\textbf{#1}}}

\newcommand{\hpad}{\hspace*{\fill}}
\newcommand{\subplotsizeslack}{1}
\NewDocumentCommand{\subplot}{ O{1} O{} m m m }{%
    \hpad%
    \begin{subfigure}{(\textwidth / #1) * \subplotsizeslack}
        \includegraphics[width=\textwidth, #2]{#3}
        \caption{#4}\label{#5}
    \end{subfigure}%
}

\ExplSyntaxOn
\RenewDocumentCommand \printorcid { } { }
\cs_set:Npn \__first_foot:
{
  \parbox[t]{\linewidth}
  {
     \rule{\linewidth}{.2pt}\\
     \__first_footerline: \hfill Page~ \thepage
  }
}
\cs_set:Npn \__cas_foot:
{
  \parbox[t]{\linewidth}
  {
   \rule{\linewidth}{.2pt}\\
   \sffamily\small
   \__first_footerline:
   \hfill Page~\thepage
  }
}
\csname ps@cas\endcsname
\ExplSyntaxOff

\begin{document}
\let\WriteBookmarks\relax
\def\floatpagepagefraction{1}
\def\textpagefraction{.001}

\shorttitle{Ontology-Compatible Definition Generation}
\shortauthors{Feng et~al.}

\title[mode=title]{
    From Fluent Text to Ontology-Compatible Definitions:
    A Quality-Control Agentic Workflow
}

\author[1,2]{Kun Feng}[orcid=0009-0003-1107-0047]

\author[3,4]{Chenchen Chu}

\author[5]{Yan Liu}

\author[2]{Haojie Sun}

\author[2]{Jijun Tang}

\author[1]{Chen Tian}

\author[6]{Shuolin Xiao}
\cormark[1]
\ead{sxiao20@jh.edu}

\author[7]{Jianping He}
\cormark[1]
\ead{hejianping@bfsu.edu.cn}

\author[2]{Zhao Li}
\cormark[1]
\ead{lizhao.informatics@gmail.com}

\affiliation[1]{
    organization={Nanjing University},
    city={Nanjing},
    postcode={210023},
    country={China}
}

\affiliation[2]{
    organization={Faculty of Computer Science and Artificial Intelligence,
    Shenzhen University of Advanced Technology},
    city={Shenzhen},
    postcode={518107},
    country={China}
}

\affiliation[3]{
    organization={Materials Artificial Intelligence Center, Shenzhen
    Institutes of Advanced Technology, Chinese Academy of Sciences},
    city={Shenzhen},
    postcode={518055},
    country={China}
}

\affiliation[4]{
    organization={University of Chinese Academy of Sciences},
    postcode={100049},
    city={Beijing},
    country={China}
}

\affiliation[5]{
    organization={Ambulatory Operations, Baylor College of Medicine},
    city={Houston},
    state={Texas},
    postcode={77030},
    country={USA}
}

\affiliation[6]{
    organization={Ralph O'Connor Sustainable Energy Institute, Johns Hopkins
    University},
    city={Baltimore},
    state={Maryland},
    postcode={21218},
    country={USA}
}

\affiliation[7]{
    organization={Beijing Foreign Studies University Artificial Intelligence
    and Human Languages Lab},
    country={China}
}

\cortext[cor1]{Corresponding authors}

\begin{abstract}
    Biomedical ontologies are structured knowledge systems that must evolve
    continuously to represent newly emerging knowledge, but their manual
    maintenance remains time-consuming and labor-intensive.  Textual
    definitions are especially costly because they must be readable to humans
    and compatible with controlled entities, relations, and logical
    constraints.  Large language models (LLMs) offer a promising way to assist
    ontology engineering tasks, including definition generation.  However,
    textual plausibility does not guarantee knowledge-system compatibility.  A
    generated definition may be fluent and close to a reference definition, yet
    still fail to ground its implied entities or violate ontology-level
    constraints.  Directly inserting such definitions into ontology libraries
    may therefore introduce entity-level hallucinations and logical
    inconsistencies.
    To address this problem, we propose two structured-knowledge
    incompatibility metrics, Entity Error Rate (EER) and Logical Inconsistency
    Rate (LIR), to quantify entity grounding and logical consistency for
    LLM-generated ontology definitions.  We further develop an agentic
    quality-control (QC) workflow that treats definition generation as an
    iterative process rather than a direct decoding step: it generates
    definitions, extracts axioms, checks entity existence and logical
    consistency, and uses the resulting diagnostics to drive repair loops.
    Comprehensive experiments on Disease Ontology, Uberon Ontology, and
    Uberon-Constraint-Stress demonstrate that EER and LIR complement
    conventional reference-based metrics by evaluating whether generated
    content can be grounded and checked inside the target ontology.  The
    workflow effectively decreases EER while maintaining strong conventional
    text-generation performance.  LIR further exposes sparse ontology-level
    logical failures that are invisible to text similarity metrics, although
    their repair remains limited.
\end{abstract}

\begin{keywords}
    Large language models \sep
    Biomedical ontologies \sep
    Structured knowledge systems \sep
    Knowledge compatibility \sep
    Quality control \sep
    Ontology engineering
\end{keywords}

\maketitle

\section{Introduction}\label{sec:intro}

Biomedical ontologies are structured knowledge systems that combine
terminological resources with formal knowledge bases
\citep{gruber1993translation,smith2007obo,world2012owl,jackson2021obo}.
A class definition is therefore more than a readable gloss.  It supports
search, annotation, curation, and reasoning over the ontology graph.  In
large Open Biological and Biomedical Ontologies (OBO)-style resources such as
the Disease Ontology (DO)
\citep{schriml2012disease,schriml2019human,schriml2022human}
and Uberon \citep{mungall2012uberon}, curators must keep textual definitions
aligned with class hierarchy, cross-ontology relationships, and logical
constraints.  Each release adds new classes, and many existing classes still
need definitions.  This creates a recurring authoring burden.

Large language models (LLMs) are a natural candidate for reducing this
burden because they can turn sparse ontology context, such as class names and
parent terms, into fluent draft definitions at scale
\citep{brown2020language,babaei2023llms4ol,val2025ontogenix}.
These drafts can reduce the
amount of manual writing curators must perform from scratch.
Recent biomedical LLM studies further show that domain-adapted language
models can support specialized scientific text tasks
\citep{labrak2024biomistral}.
Given a term name and its parent context, an LLM can often produce fluent
biomedical definitions that resemble curated ontology text.  This makes
LLM-assisted definition generation a plausible way to support ontology
maintenance.

However, LLM-generated drafts still require ontology-aware validation, and
the main challenge is evaluation.
Standard generation metrics quantify whether a generated definition is
lexically or semantically close to a reference.  These metrics include token
overlap, Recall-Oriented Understudy for Gisting Evaluation (ROUGE),
Bilingual Evaluation Understudy (BLEU), BERTScore, a metric based on
Bidirectional Encoder Representations from Transformers (BERT), and
sentence-embedding similarity
\citep{papineni2002bleu,lin2004rouge,zhang2019bertscore,reimers2019sentence}.
These metrics are useful, but they measure text similarity rather than
compatibility with the target knowledge system.  A definition can be fluent
and reference-like while still mentioning a nonexistent entity, using an
unresolved relation, or adding a relationship that makes the class
unsatisfiable after it is merged into the ontology.

\begin{figure*}[t]
    \centering
    \includegraphics[width=\linewidth]{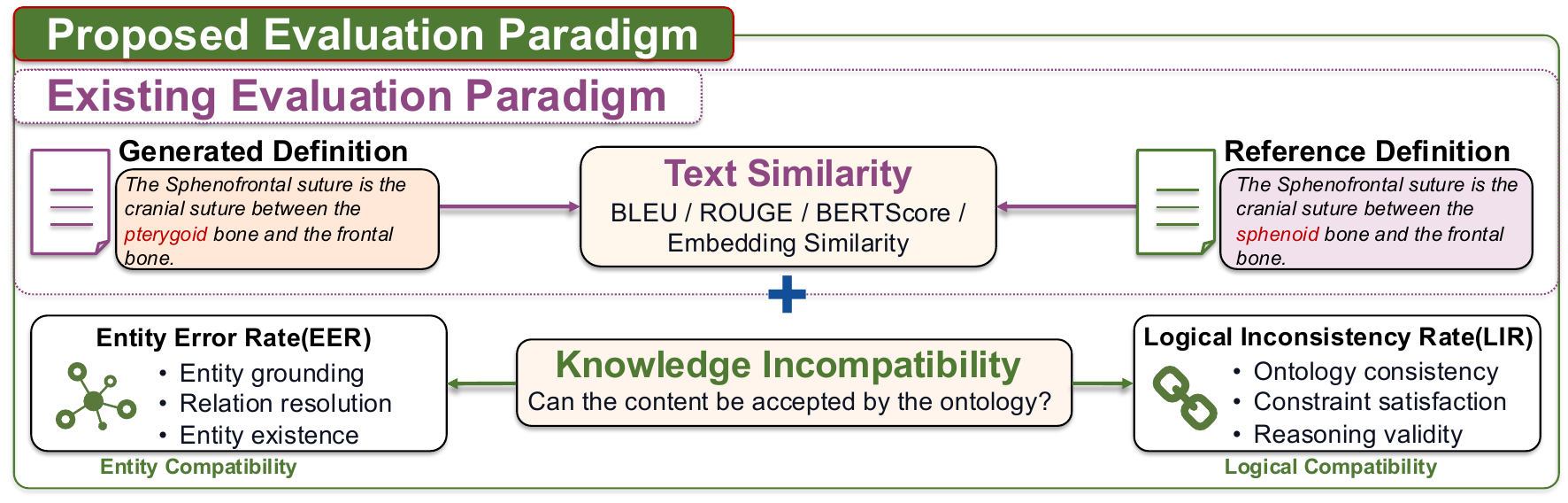}
    \caption{Evaluation framework for LLM-generated ontology definitions.
    Text-generation metrics measure similarity to reference definitions.
    Knowledge-compatible evaluation, instantiated here for ontologies, also
    checks entity grounding and logical consistency.}
    \label{fig:intro:evaluation-framework}
\end{figure*}

\Cref{fig:intro:evaluation-framework} illustrates this gap with a concrete
definition example.  The generated definition is fluent and closely resembles
the reference definition, but it changes the anatomical entity from
``sphenoid bone'' to ``pterygoid bone.''  A text-similarity metric may still
assign a high score because most words and sentence structure are shared.
For ontology use, however, this is not enough.  The mentioned entity must be
grounded to an identifier in the target ontology, the implied relation must be
resolved, and the resulting axiom must be compatible with existing ontology
constraints.  If the entity cannot be grounded
or an axiom conflicts with disjointness or other strong
constraints, it can make the generated class unsatisfiable
\citep{ji2023survey}.

We therefore study ontology definition generation as a quality-controlled
knowledge-compatible generation problem.  Instead of treating the generated
definition as the final artifact, we place generation inside an agentic
quality-control (QC) workflow.  The workflow uses ontology tools to inspect
intermediate outputs, receive diagnostic feedback, and repair candidates when
needed.  This design follows a broader line of LLM agent and tool-use work in
which models call external tools, use feedback, and refine outputs over
multiple steps
\citep{yao2022react,schick2023toolformer,madaan2023self,shinn2023reflexion}.

We also introduce two metrics that separate structured-knowledge failures
from ordinary text quality.  Entity Error Rate (EER) measures unresolved or
ungrounded extracted axioms.  Logical Inconsistency Rate (LIR) measures
reasoner-detected incompatibility among entity-grounded candidates.  These
metrics are instantiated here for biomedical ontologies, but they reflect a
broader principle: generated content for a structured knowledge system should
be checked against the entities, relations, and constraints of that system.

Our evaluation covers both realistic release-based ontology maintenance
settings and a targeted stress setting.  For DO and Uberon, we evaluate on
incremental release data that reflect newly added or newly defined classes.
To better assess logical consistency, we also construct
Uberon-Constraint-Stress, a subset enriched for classes involved in strong
constraints such as disjointness.  Together, these settings test whether the
QC workflow can detect entity-grounding and logical failures that ordinary
text metrics cannot.

This paper makes four contributions:
\begin{itemize}
    \item We frame LLM-assisted ontology definition generation as a
    knowledge-compatibility problem, instantiated in biomedical ontologies,
    rather than only as a text-generation problem.
    \item We propose two structured-knowledge incompatibility metrics: Entity
    Error Rate (EER), which measures entity-grounding failures, and Logical
    Inconsistency Rate (LIR), which measures reasoner-detected incompatibility
    among entity-grounded candidates.
    \item We introduce an axiom-based agentic QC workflow that combines
    definition generation, axiom extraction, entity existence checking,
    ROBOT-based reasoning, and iterative repair.
    \item We evaluate the workflow on DO, Uberon, and
    Uberon-Constraint-Stress, report both conventional text-generation metrics
    and structured-knowledge incompatibility metrics, and release the constructed
    evaluation datasets for reuse.
\end{itemize}
Overall, the results show that fluent, reference-like definitions are not
necessarily ready for integration into a target knowledge system.  The QC
modules expose two failure modes that text metrics miss: extracted entities or
relations may fail to ground, and grounded candidates may still conflict with
ontology-level constraints.  The repair loops reduce many entity-grounding
failures before curator review, while the remaining logical failures show why
symbolic validation is still needed even when generated text appears strong.
More broadly, these findings support a layered view of LLM-assisted
structured-knowledge generation: language models can draft useful content, but
external knowledge-system checks should decide whether that content can be
grounded, repaired, or safely accepted.

\section{Related Work}\label{sec:related}

\subsection{Biomedical Ontologies and Definition Curation}

Biomedical ontologies provide shared class identifiers, labels, definitions,
relations, and logical axioms for data integration.  This follows the broader
view of an ontology as an explicit specification of a conceptualization
\citep{gruber1993translation}.  The OBO Foundry
established principles for coordinated ontology development, including
openness, common formats, orthogonality, textual definitions, and formal
relations \citep{smith2007obo,jackson2021obo}.  The Web Ontology Language
(OWL) provides a formal language for representing classes, relations, and
logical axioms in machine-interpretable ontologies \citep{world2012owl}.  DO
provides a structured disease
vocabulary for integrating disease knowledge across biomedical resources
\citep{schriml2012disease,schriml2019human}.  Uberon provides a cross-species
anatomy ontology that integrates anatomical entities and supports reasoning
through rich logical relationships
\citep{mungall2012uberon}.

Definition authoring in these resources is labor-intensive because a useful
definition must be readable to domain users while remaining compatible with
the ontology's formal structure.  Missing definitions limit interpretability,
and weak or inconsistent definitions reduce the utility of ontology-mediated
search, annotation, and reasoning.  Our work focuses on this curation
bottleneck, but treats definition generation as one step in a broader
validation workflow rather than as standalone text production.  This focus is
consistent with ontology engineering practices that emphasize reuse,
cross-resource lookup, and automated release workflows
\citep{courtot2011mireot,whetzel2011bioportal,matentzoglu2022ontology}.

\subsection{LLMs for Ontology Engineering}

Recent work has explored LLMs for ontology learning and ontology engineering,
including term typing, taxonomy discovery, non-taxonomic relation extraction,
ontology population, and ontology generation
\citep{brown2020language,babaei2023llms4ol,zhang2024ontochat,kommineni2024human,fathallah2024neon,li2025large,val2025ontogenix}.
These
studies show that LLMs can provide useful candidate structures and reduce
manual modeling effort, while also observing that complex modeling decisions
remain difficult and require validation.  Parameter-efficient adaptation
methods such as Low-Rank Adaptation (LoRA) make it practical to specialize
open LLMs for domain-specific generation tasks \citep{hu2022lora,qwen2.5}.
Definition generation is closely related: the model must synthesize fluent
domain text from ontology context, and the generated text may contain implicit
ontological commitments that affect the formal model.

Biomedical representation learning has also used definitions as an important
source of semantic grounding.  For example, definition-based biomedical
concept representations can better capture concept meaning than names alone
\citep{remy2022biolord}.  Our goal is complementary.  Rather than using
definitions only as training signal for representations, we generate candidate
definitions and then test whether they can be grounded back into ontology
entities and axioms.  This is also related to schema-guided extraction work
such as SPIRES/OntoGPT, where LLM outputs are normalized against existing
ontologies and vocabularies \citep{caufield2024structured}.

\subsection{Text- and Knowledge-Level Evaluation of Generated Content}

Automatic text-generation metrics such as BLEU, ROUGE, BERTScore, and
embedding cosine similarity are useful for measuring lexical overlap,
sequence-level similarity, and contextual semantic similarity
\citep{papineni2002bleu,lin2004rouge,zhang2019bertscore,reimers2019sentence}.
For generated ontology definitions, these metrics are useful baselines
because a candidate definition should generally resemble curated text for the
same class.  However, text-level similarity does not show whether the content
can be grounded in, checked by, and integrated into the target knowledge
system.

Recent work on factuality and hallucination evaluation has moved in this
direction.  Instead of judging only whether generated text sounds plausible or
matches a reference, these approaches decompose text into checkable claims or
compare generated content with external knowledge resources
\citep{ji2023survey,min2023factscore}.  Related work on LLMs and
knowledge graphs also emphasizes that structured knowledge can be used to
ground model outputs and identify unsupported claims
\citep{pan2024unifying,lavrinovics2025knowledge}.  This line of work is
closer to the needs of ontology generation than surface text metrics alone,
because ontology definitions are connected to controlled entities, relations,
and formal constraints.  The remaining gap is system-level compatibility:
even a plausible statement must use entities and relations that the target
knowledge system can resolve and validate.

Ontology definition generation adds a further requirement because ontology
libraries are structured knowledge systems.  A new term should be explained
using existing ontology terms, rather than free text alone.  Terms are
connected through subclass, partonomy, disjointness, and other relations, and
each ontology term is associated with at least one axiom.  Thus, generated
definitions may imply relationship axioms that must be usable inside the
ontology.  A valid-looking definition can still mention an entity that does
not exist, use a relation that cannot be resolved, or produce an axiom that
makes a class unsatisfiable.  Therefore, knowledge-compatible evaluation needs
signals beyond reference similarity.  In our workflow, text metrics remain
part of the evaluation, but they are complemented by axiom parse rate,
predicate--object matching, entity grounding checks, EER, and LIR.  This
combination exposes cases where two definitions are textually similar but
ontologically different, or where a fluent definition cannot be safely merged
into the ontology.

\subsection{Ontology Quality Control and Reasoning}

Ontology engineering workflows commonly use automated tools to enforce
format, merge, and reasoning checks.  ROBOT is a standard command-line tool
for automating ontology workflows, including template conversion, merging,
module extraction, and reasoning \citep{jackson2019robot}.  Reasoners such as
ELK, an OWL 2 EL reasoner, are widely used for scalable OWL EL classification
and satisfiability checking \citep{kazakov2014incredible}.  These tools are
typically applied to curated ontology releases or manually edited ontology
modules.

Our work brings these QC tools into the LLM generation loop.  Generated text
is not accepted solely because it is fluent or close to a reference.  Instead,
it is converted into candidate axioms, checked for resolvable entities,
merged into a replacement ontology, and tested with ROBOT reasoning.  This
turns knowledge-compatible generation into an operational workflow for
ontologies: LLMs propose, symbolic checks diagnose, and repair loops revise
the outputs before they reach human curation.

\section{Methodology}\label{sec:method}

\begin{figure*}[t]
    \centering
    \includegraphics[width=\linewidth]{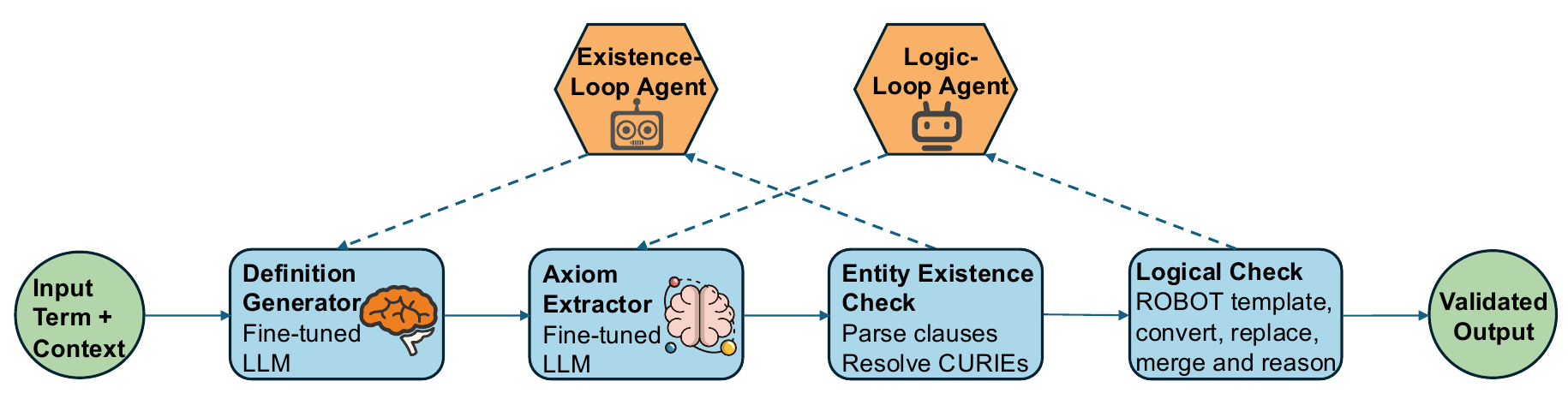}
    \caption{Axiom-based quality-control workflow.  The existence loop retries
    definition generation and axiom extraction for records with unresolved
    predicates or objects.  The logic loop keeps the generated definition fixed
    and revises only the extracted axioms for records that become
    unsatisfiable after ROBOT reasoning; it does not trigger definition
    regeneration, because allowing both definition-level and axiom-level
    repair in the logic loop could create many additional iterations even
    with per-loop retry limits.}\label{fig:method:workflow}
\end{figure*}

We formulate ontology definition generation as a constrained
knowledge-compatible generation problem.  Biomedical ontologies provide a
concrete setting in which generated text can be checked against the entities,
relations, and constraints of a structured knowledge system.

\Cref{fig:method:workflow} gives an overview of the proposed agentic QC
workflow.  For each ontology class, the input contains the class name and
parent context.  The workflow first generates a candidate textual definition.
It then extracts predicate--object relationship axioms from that definition,
so the generated text can be tested as structured ontology content.  The
extracted predicates and objects are resolved against the target ontology.  If
they can be grounded, the workflow converts the candidate into a ROBOT
template and validates it with an ontology reasoner
\citep{jackson2019robot,kazakov2014incredible}.

The workflow uses these tool outputs as feedback for targeted repair.
Candidates with unresolved predicates or objects enter an existence loop,
which regenerates the definition and re-extracts axioms.  Candidates that are
grounded but make the ontology unsatisfiable enter a logic loop, which keeps
the generated definition fixed and requests axiom-only correction.  A
candidate is accepted only after it passes the relevant grounding and
reasoning checks.  The same workflow structure is used for DO and Uberon,
with ontology-specific configuration for prefixes, relation labels, release
files, and local lookup indices.

\subsection{Task Formulation}\label{sub:method:task}

Each input record contains a class name and its parent classes.  Parent
context is represented by parent names and parent textual definitions.  For
supervised training and evaluation, each record also contains a reference
definition and a set of relationship axioms.  We use two generation stages.
The first stage produces a draft textual definition from the class name and
parent context.  The second stage reads that draft and extracts explicit
predicate--object relationship axioms, which provide the structured content
used by the downstream grounding and reasoning checks.

Let \(x_i\) denote the \(i\)-th input record, consisting of the target class
name and its parent context.  The definition generator maps \(x_i\) to a
candidate textual definition \(d_i\).  The axiom extractor then reads
\(d_i\), together with the same class name and parent context, and returns a
set of extracted relationship axioms for that record:
\[
    A_i = \{a_{ij}\}_{j=1}^{m_i}, \qquad
    a_{ij} = (p_{ij}, o_{ij}).
\]
Here \(m_i\) is the number of extracted axioms for record \(i\), \(j\) indexes
one extracted axiom within that record, \(p_{ij}\) is the relation predicate,
and \(o_{ij}\) is the object entity.  For example, an extracted axiom may
state that the target class \emph{part of} some anatomical structure, where
\emph{part of} is the predicate and the anatomical structure is the object.
The extractor emits axioms in a
controlled textual format:
\begin{quote}
\small
\texttt{relationship: predicate: [PREDICATE\_NAME], object: [OBJECT\_NAME];}
\end{quote}
This intermediate format allows the workflow to parse model output, resolve
predicate and object names into ontology identifiers, and test whether the
generated definition can be safely integrated into the ontology.

\subsection{Dataset Construction}\label{sub:method:datasets}

The datasets are built from ontology GitHub release files.  For each release, we
collect active, non-obsolete ontology classes, their names, textual
definitions, parent classes, and relationship axioms.  Definition-generation
records use the class name and parent context as input and the curated
definition as the target.  Axiom-extraction records use the class name,
definition, and parent context as input and the ontology-provided
relationship axioms as the target.

We first use DO as an initial testbed for evaluating whether generated
relationship axioms remain valid after being merged into a complete ontology.
For DO, we construct a 2023 training set and two release-based test sets:
2024 and 2025.  A DO test record is included when a class is
newly added or newly receives a definition in the later release, and when it
has relationship axioms for downstream validation.  However, strong logical
constraints are sparse in DO: in the 2023 release, only 26 of 11,454 active DO
terms are involved in strong constraint axioms such as \texttt{DisjointFrom}.
This makes DO useful for entity-grounding evaluation but less suitable for
stress-testing logical inconsistency after ontology merging.

We therefore add Uberon as a second evaluation ontology.  In addition to
release-based Uberon test records from newly added or newly defined terms in
the 2024 and 2025 releases, we construct Uberon-Constraint-Stress from the
2023 release.  This stress set contains all 973 terms involved in strong
constraint axioms, and the remaining 9,557 terms are used as the main Uberon
training pool.  This separation keeps constraint-focused terms out of the main
fine-tuning data, so Uberon-Constraint-Stress provides a targeted test of
knowledge-compatibility failures inside an ontology.

\subsection{Definition Generation}\label{sub:method:defgen}

The definition generation stage uses a chat-style LLM prompt.  The system
prompt instructs the model to act as a biomedical ontology expert and produce
concise ontology-style definitions.  The user prompt contains only the target
term name and parent context.  In the initial pass, the model is asked to
return only the generated definition.  In existence-repair passes, the prompt
also includes feedback from the previous failed attempt, including the
previous generated definition and unresolved predicate or object diagnostics.

The implementation uses two model families, Qwen2.5-14B-Instruct
\citep{qwen2.5} and Llama-3.1-8B-Instruct
\citep{dubey2024llama}, each evaluated in a base setting and a
LoRA-adapted setting fine-tuned on ontology definition records
\citep{hu2022lora}.  The fine-tuning and random-sampling repair settings are
reported in \Cref{tab:app:hyperparameters}.  Generated text is stored with
the original ontology metadata for downstream axiom extraction.

\subsection{Relationship Axiom Extraction}\label{sub:method:axiom}

The axiom extraction stage converts a generated definition into explicit
relationship axioms.  Its prompt contains the term name, generated definition,
and parent context, and asks the model to return one or more axiom lines in
the controlled predicate-object format.  The model uses predicate names and
object names rather than Compact Uniform Resource Identifiers (CURIEs) or
Internationalized Resource Identifiers (IRIs), because the next stage resolves
these names against the target ontology.  For each model family, the base
setting uses a five-shot prompting setup for axiom extraction.  The five
examples are fixed training-set records selected to cover terms with different
numbers of relationship axioms, so that the prompt includes both simple and
axiom-rich cases.  The fine-tuned setting replaces this few-shot prompting
setup with the corresponding LoRA-adapted axiom extractor trained on ontology
axiom records.

For evaluation, reference axioms are attached to definition records by
matching ontology identifiers against the axiom dataset.  This enables
standard axiom-extraction metrics such as parse rate, exact match,
predicate-object pair precision/recall/F1, predicate accuracy, object
accuracy, and top-1 accuracy.  These metrics evaluate whether the generated
axiom text matches ontology-provided axioms, but they do not guarantee that
the generated output is valid in the target ontology.

\subsection{Entity Existence Checking and EER}\label{sub:method:existence}

The first QC module checks whether each extracted axiom can be grounded in the
target knowledge system.  In the ontology setting, this means that the
workflow parses the controlled axiom text into predicate-object pairs, resolves
each predicate to a relationship property, and resolves each object to an
ontology entity.  Resolution first uses a local ontology index built from the
target OBO file; Ontology Lookup Service 4 (OLS4) lookup can be enabled for
external label resolution \citep{cote2006ontology}.  The check records
several failure types, including unparsed axiom text, unresolved predicates,
and unresolved objects.

A record passes the existence check only if at least one axiom is parsed and
all parsed predicate-object pairs can be resolved.  Resolved axioms are
serialized into CURIE-based form and passed to ROBOT.  We define EER as
\begin{equation}
    \mathrm{EER}
    =
    \frac{1}{N}
    \sum_{i=1}^{N}
    \mathbb{I}\{\mathrm{ExistencePass}(A_i) = 0\},
    \label{eq:method:eer}
\end{equation}
where \(N\) is the number of generated definition records and
\(\mathbb{I}\{\cdot\}\) is an indicator function.  EER therefore measures the proportion of
generated records whose extracted axioms cannot be grounded into the target
knowledge system, instantiated here as the target ontology.

\subsection{Logical Consistency Checking and LIR}\label{sub:method:logic}

Records that pass entity existence checking are then tested against the
constraints of the target knowledge system.  In our ontology instantiation,
this validation is performed with ROBOT \citep{jackson2019robot}.  The
workflow creates a dynamic ROBOT template comma-separated values (CSV) file
whose rows contain the class identifier (ID), label, generated definition,
class type, and one existential restriction column for each extracted
predicate.  A predicate-object axiom is rendered as
a subclass restriction of the form \texttt{SC P some \%}.  Before merging, the
workflow writes the generated class IDs to a replacement-term file and removes
their existing records from the base ontology.  This replacement step prevents
candidate axioms from being combined with older axioms for the same classes.
The template ontology is then merged with the pruned base ontology and checked
with a reasoner through ROBOT.  HermiT provides more complete
reasoning support than ELK, but its memory requirements were prohibitive in
our setting: even with 400 GB of allocated memory and a ROBOT-extract strategy
that reasons over local modules, HermiT did not complete successfully.
Therefore, the final workflow uses ELK for reasoning
\citep{kazakov2014incredible,glimm2014hermit}.

ROBOT reports classes that become unsatisfiable after reasoning.  We count
only unsatisfiable classes from the generated candidate set and ignore
unrelated unsatisfiable classes already present in the ontology.  If a
candidate causes a ROBOT Manchester syntax parse error, the workflow removes
that record, records it as a template parse skip, and reruns reasoning on the
remaining candidates.

We define LIR as
\begin{equation}
    \mathrm{LIR}
    =
    \frac{1}{N_{\mathrm{exist}}}
    \sum_{i=1}^{N_{\mathrm{exist}}}
    \mathbb{I}\{c_i \in U_{\mathrm{workflow}}\},
    \label{eq:method:lir}
\end{equation}
where \(U_{\mathrm{workflow}}\) is the set of workflow candidate classes
reported as unsatisfiable by ROBOT reasoning, and \(N_{\mathrm{exist}}\) is
the number of candidates that passed entity existence checking and entered the
logic check.  The term \(\mathbb{I}\{\cdot\}\) is the same indicator function
defined above.  LIR measures ontology-level incompatibility among candidates
that are already entity-grounded.

\subsection{Iterative Repair Loops}\label{sub:method:loops}

The repair procedure gives the workflow its agentic character: each repair
step conditions generation on tool-produced diagnostics, keeps the failed
candidate and retry history as context, and chooses the next action according
to the type of failure.  This agentic repair is implemented through two loops
with different scopes.  The existence loop handles unresolved predicates,
unresolved objects, ambiguous labels, and
unparsed axiom text.  For each failed record, the retry prompt keeps the
original term metadata, the previous generated definition, and the
existence-failure diagnostics.  The definition generator then produces a new
definition, the axiom extractor extracts new axioms, and the existence checker
is rerun.  Passing records are accepted into the candidate set.  Records that
remain unresolved are retried until the loop budget is exhausted.

The logic loop handles candidates that pass entity existence checking but
become unsatisfiable after ROBOT reasoning.  In this loop, the textual
definition is kept fixed.  Only the relationship axioms are revised.  The
logic-failed record is enriched with feedback stating that the class was
reported unsatisfiable, the current extracted axioms, and a compact history of
previous failed axiom sets.  The axiom correction prompt instructs the model
to correct unsupported axioms, especially those that may contradict
taxon, disjointness, partonomy, or domain/range constraints.  Corrected axioms
must pass the existence check before they replace the previous candidate
axioms.  The logic loop then reruns ROBOT reasoning until no workflow
candidates are unsatisfiable or the loop budget is exhausted.

\subsection{Evaluation Metrics}\label{sub:method:metrics}

We report both conventional text-generation metrics and structured-knowledge
QC metrics.  For generated definitions, the code computes token F1, ROUGE-L,
BLEU-2, BERTScore F1 (reported as BERT-F1 in tables), and
sentence-embedding cosine similarity against curated reference definitions
\citep{papineni2002bleu,lin2004rouge,zhang2019bertscore,reimers2019sentence}.
BLEU-2 is computed as the average of 1--2-gram BLEU scores.  We use BLEU-2
rather than 3--4-gram BLEU because most DO definitions are relatively short,
making longer n-gram overlap sparse and less stable.
For axiom extraction, it computes parse rate, exact match, predicate-object
pair precision/recall/F1, predicate accuracy, object accuracy, and top-1
accuracy against ontology-provided relationship axioms.

The two primary QC metrics are EER and LIR.  EER measures failures of entity
grounding, while LIR measures reasoner-detected inconsistency among
entity-grounded candidates.  Together, these metrics capture ontology
compatibility errors that may not be reflected by reference-based text
similarity metrics.

\section{Experimental Results}\label{sec:results}

\subsection{Experimental Setup}\label{sec:results:setup}

\begin{figure*}[t]
    \centering
    \includegraphics[width=\linewidth]{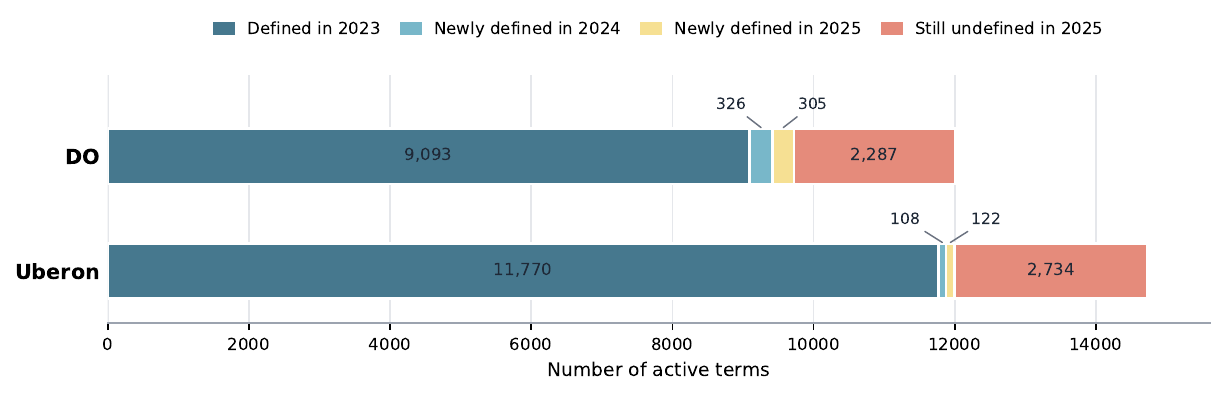}
    \caption{Definition coverage and remaining definition gaps in DO and
    Uberon.  The 2024 and 2025 segments combine newly added active terms with
    definitions and previously active terms that first received definitions in
    the newer release.}
    \label{fig:results:definition-coverage}
\end{figure*}

We evaluate the proposed workflow on release-based ontology maintenance tasks
and on a constraint-focused stress test.  The release-based tasks simulate a
curation scenario in which a model is trained on an older ontology release and
evaluated on terms that are newly added or newly defined in a later release.
We use this setup for both Disease Ontology (DO) and Uberon.  DO provides an
initial testbed for release-based definition generation, but it contains too
few strong constraints for stress-testing logical inconsistency: in the 2023
release, only 26 of 11,454 active DO terms are involved in strong constraint
axioms such as \texttt{DisjointFrom}.  We therefore include Uberon and
Uberon-Constraint-Stress, a targeted subset of Uberon classes involved in
strong constraints.

All release-based splits are constructed from GitHub release snapshots.  This
choice reflects the practical maintenance setting shown in
\Cref{fig:results:definition-coverage}: both DO and Uberon still contain many
active terms without definitions in 2025, while each annual release adds
definitions for only a relatively small number of terms.  The DO training set
uses the December 2023 release.  DO 2024 contains terms newly added or newly
defined between the December 2023 and December 2024 releases, and DO 2025 uses
the same criterion between the December 2024 and December 2025 releases.  For
Uberon, the training set uses the 2023-12-08 release.  Uberon 2024 compares the
2023-12-08 and 2024-11-25 releases, and Uberon 2025 compares the 2024-11-25 and
2025-12-05 releases.  The stress subset is held out from the main Uberon
training pool and is used to make ontology-level failures easier to observe.
For both ontologies, the training pool is split into training and validation
subsets with an 8:2 ratio.  \Cref{tab:result:datasets} summarizes the
resulting train/validation sizes and evaluation sets.

\begin{table*}[t]
    \centering
    \caption{Training and evaluation settings used in the main experiments.}
    \label{tab:result:datasets}
    \small
    \begin{tabular}{llccc}
        \toprule
        Ontology & Test setting & Definition train+val & Axiom train+val & Test size \\
        \midrule
        DO & 2024 & 9,093 & 4,810 & 188 \\
        DO & 2025 & 9,093 & 4,810 & 236 \\
        Uberon & 2024 & 9,557 & 9,557 & 96 \\
        Uberon & 2025 & 9,557 & 9,557 & 112 \\
        Uberon & Constraint-Stress & 9,557 & 9,557 & 973 \\
        \bottomrule
    \end{tabular}
\end{table*}

We compare Qwen2.5-14B and Llama-3.1-8B under base and LoRA fine-tuned
settings.  All reasoning-based checks use ELK, and entity-existence validation
uses the local ontology together with OLS4 lookup.  When a repair loop is
enabled, the workflow allows at most five retry iterations; no-loop baselines
run direct generation once.  Retry generations use random sampling so that
each repair iteration can explore an alternative candidate rather than simply
repeat the previous output.  For each setting, we report conventional
reference-based text metrics and two structured-knowledge incompatibility
metrics: EER and LIR.

\subsection{Entity Error Rate Results}\label{sec:results:main}

\Cref{tab:result:main} places EER next to conventional text-generation
metrics for the same no-loop and existence-loop pairs.  It shows that models
can achieve reasonable conventional text-generation scores while still having
high EER, especially in base-model settings.  After the existence loop, the
conventional metric columns usually retain similar shading within each pair,
indicating that the repair loop maintains strong surface-similarity
performance.  In contrast, the EER column often becomes darker, because lower
EER is better.  This pattern supports the interpretation that the main benefit
of the loop is improved ontology compatibility while preserving
reference-based text quality.

\begin{table*}[t]
    \centering
    \scriptsize
    \caption{Conventional text-generation scores and Entity Error Rate
    (EER) for the same model-loop settings shown in \Cref{fig:result:eer}.
    Darker cells indicate better values: higher conventional scores and
    lower EER.  Arrows in column headers indicate the preferred
    metric direction.}
    \label{tab:result:main}
    \setlength{\tabcolsep}{3pt}
    \resizebox{\linewidth}{!}{%
    \begin{tabular}{lllcccccc}
        \toprule
        Test setting & Model & Loop setting & Token F1 \(\uparrow\) & ROUGE-L \(\uparrow\) & BLEU-2 \(\uparrow\) & BERT-F1 \(\uparrow\) & Cosine \(\uparrow\) & EER \(\downarrow\) \\
        \midrule
        DO 2024 & Qwen2.5-14B & Base, no loop & \cellcolor{MetricBlue!28}0.283 & \cellcolor{MetricBlue!23}0.222 & \cellcolor{MetricBlue!14}0.107 & \cellcolor{MetricBlue!54}0.608 & \cellcolor{MetricBlue!58}0.664 & \cellcolor{MetricBlue!31}0.670 \\
        DO 2024 & Qwen2.5-14B & Base, existence loop & \cellcolor{MetricBlue!28}0.293 & \cellcolor{MetricBlue!24}0.236 & \cellcolor{MetricBlue!14}0.118 & \cellcolor{MetricBlue!54}0.619 & \cellcolor{MetricBlue!59}0.676 & \cellcolor{MetricBlue!55}0.378 \\
        DO 2024 & Qwen2.5-14B & LoRA, no loop & \cellcolor{MetricBlue!52}0.588 & \cellcolor{MetricBlue!51}0.579 & \cellcolor{MetricBlue!35}0.380 & \cellcolor{MetricBlue!63}0.731 & \cellcolor{MetricBlue!65}0.766 & \cellcolor{MetricBlue!65}0.016 \\
        DO 2024 & Qwen2.5-14B & LoRA, existence loop & \cellcolor{MetricBlue!52}0.588 & \cellcolor{MetricBlue!51}0.579 & \cellcolor{MetricBlue!35}0.381 & \cellcolor{MetricBlue!63}0.731 & \cellcolor{MetricBlue!65}0.766 & \cellcolor{MetricBlue!65}0.000 \\
        DO 2024 & Llama-3.1-8B & Base, no loop & \cellcolor{MetricBlue!30}0.310 & \cellcolor{MetricBlue!24}0.240 & \cellcolor{MetricBlue!14}0.115 & \cellcolor{MetricBlue!52}0.593 & \cellcolor{MetricBlue!56}0.642 & \cellcolor{MetricBlue!29}0.697 \\
        DO 2024 & Llama-3.1-8B & Base, existence loop & \cellcolor{MetricBlue!31}0.320 & \cellcolor{MetricBlue!25}0.244 & \cellcolor{MetricBlue!15}0.123 & \cellcolor{MetricBlue!54}0.608 & \cellcolor{MetricBlue!58}0.656 & \cellcolor{MetricBlue!48}0.457 \\
        DO 2024 & Llama-3.1-8B & LoRA, no loop & \cellcolor{MetricBlue!51}0.580 & \cellcolor{MetricBlue!50}0.562 & \cellcolor{MetricBlue!35}0.372 & \cellcolor{MetricBlue!63}0.725 & \cellcolor{MetricBlue!65}0.764 & \cellcolor{MetricBlue!65}0.005 \\
        DO 2024 & Llama-3.1-8B & LoRA, existence loop & \cellcolor{MetricBlue!51}0.580 & \cellcolor{MetricBlue!50}0.562 & \cellcolor{MetricBlue!35}0.372 & \cellcolor{MetricBlue!63}0.725 & \cellcolor{MetricBlue!65}0.764 & \cellcolor{MetricBlue!65}0.005 \\
        \midrule
        DO 2025 & Qwen2.5-14B & Base, no loop & \cellcolor{MetricBlue!31}0.327 & \cellcolor{MetricBlue!25}0.253 & \cellcolor{MetricBlue!16}0.143 & \cellcolor{MetricBlue!56}0.636 & \cellcolor{MetricBlue!60}0.688 & \cellcolor{MetricBlue!41}0.551 \\
        DO 2025 & Qwen2.5-14B & Base, existence loop & \cellcolor{MetricBlue!32}0.331 & \cellcolor{MetricBlue!26}0.260 & \cellcolor{MetricBlue!17}0.148 & \cellcolor{MetricBlue!56}0.636 & \cellcolor{MetricBlue!60}0.687 & \cellcolor{MetricBlue!60}0.318 \\
        DO 2025 & Qwen2.5-14B & LoRA, no loop & \cellcolor{MetricBlue!53}0.606 & \cellcolor{MetricBlue!53}0.602 & \cellcolor{MetricBlue!37}0.399 & \cellcolor{MetricBlue!64}0.739 & \cellcolor{MetricBlue!65}0.771 & \cellcolor{MetricBlue!65}0.000 \\
        DO 2025 & Qwen2.5-14B & LoRA, existence loop & \cellcolor{MetricBlue!53}0.606 & \cellcolor{MetricBlue!53}0.602 & \cellcolor{MetricBlue!37}0.399 & \cellcolor{MetricBlue!64}0.739 & \cellcolor{MetricBlue!65}0.771 & \cellcolor{MetricBlue!65}0.000 \\
        DO 2025 & Llama-3.1-8B & Base, no loop & \cellcolor{MetricBlue!33}0.345 & \cellcolor{MetricBlue!26}0.257 & \cellcolor{MetricBlue!16}0.139 & \cellcolor{MetricBlue!54}0.611 & \cellcolor{MetricBlue!58}0.658 & \cellcolor{MetricBlue!28}0.708 \\
        DO 2025 & Llama-3.1-8B & Base, existence loop & \cellcolor{MetricBlue!33}0.348 & \cellcolor{MetricBlue!26}0.260 & \cellcolor{MetricBlue!17}0.144 & \cellcolor{MetricBlue!54}0.615 & \cellcolor{MetricBlue!58}0.659 & \cellcolor{MetricBlue!49}0.449 \\
        DO 2025 & Llama-3.1-8B & LoRA, no loop & \cellcolor{MetricBlue!54}0.608 & \cellcolor{MetricBlue!53}0.605 & \cellcolor{MetricBlue!37}0.396 & \cellcolor{MetricBlue!63}0.727 & \cellcolor{MetricBlue!65}0.759 & \cellcolor{MetricBlue!65}0.017 \\
        DO 2025 & Llama-3.1-8B & LoRA, existence loop & \cellcolor{MetricBlue!54}0.608 & \cellcolor{MetricBlue!53}0.605 & \cellcolor{MetricBlue!37}0.396 & \cellcolor{MetricBlue!63}0.727 & \cellcolor{MetricBlue!65}0.759 & \cellcolor{MetricBlue!65}0.000 \\
        \midrule
        Uberon 2024 & Qwen2.5-14B & Base, no loop & \cellcolor{MetricBlue!39}0.428 & \cellcolor{MetricBlue!33}0.353 & \cellcolor{MetricBlue!21}0.203 & \cellcolor{MetricBlue!58}0.663 & \cellcolor{MetricBlue!62}0.717 & \cellcolor{MetricBlue!65}0.229 \\
        Uberon 2024 & Qwen2.5-14B & Base, existence loop & \cellcolor{MetricBlue!39}0.425 & \cellcolor{MetricBlue!33}0.351 & \cellcolor{MetricBlue!21}0.200 & \cellcolor{MetricBlue!58}0.664 & \cellcolor{MetricBlue!62}0.717 & \cellcolor{MetricBlue!65}0.010 \\
        Uberon 2024 & Qwen2.5-14B & LoRA, no loop & \cellcolor{MetricBlue!39}0.428 & \cellcolor{MetricBlue!35}0.374 & \cellcolor{MetricBlue!23}0.228 & \cellcolor{MetricBlue!61}0.699 & \cellcolor{MetricBlue!64}0.742 & \cellcolor{MetricBlue!65}0.052 \\
        Uberon 2024 & Qwen2.5-14B & LoRA, existence loop & \cellcolor{MetricBlue!39}0.431 & \cellcolor{MetricBlue!35}0.378 & \cellcolor{MetricBlue!23}0.230 & \cellcolor{MetricBlue!61}0.699 & \cellcolor{MetricBlue!64}0.742 & \cellcolor{MetricBlue!65}0.000 \\
        Uberon 2024 & Llama-3.1-8B & Base, no loop & \cellcolor{MetricBlue!38}0.409 & \cellcolor{MetricBlue!32}0.332 & \cellcolor{MetricBlue!19}0.175 & \cellcolor{MetricBlue!57}0.649 & \cellcolor{MetricBlue!61}0.698 & \cellcolor{MetricBlue!65}0.177 \\
        Uberon 2024 & Llama-3.1-8B & Base, existence loop & \cellcolor{MetricBlue!38}0.409 & \cellcolor{MetricBlue!31}0.331 & \cellcolor{MetricBlue!19}0.176 & \cellcolor{MetricBlue!57}0.650 & \cellcolor{MetricBlue!61}0.698 & \cellcolor{MetricBlue!65}0.042 \\
        Uberon 2024 & Llama-3.1-8B & LoRA, no loop & \cellcolor{MetricBlue!43}0.478 & \cellcolor{MetricBlue!39}0.421 & \cellcolor{MetricBlue!25}0.249 & \cellcolor{MetricBlue!61}0.695 & \cellcolor{MetricBlue!64}0.735 & \cellcolor{MetricBlue!65}0.219 \\
        Uberon 2024 & Llama-3.1-8B & LoRA, existence loop & \cellcolor{MetricBlue!44}0.484 & \cellcolor{MetricBlue!39}0.424 & \cellcolor{MetricBlue!26}0.257 & \cellcolor{MetricBlue!61}0.701 & \cellcolor{MetricBlue!64}0.743 & \cellcolor{MetricBlue!65}0.052 \\
        \midrule
        Uberon 2025 & Qwen2.5-14B & Base, no loop & \cellcolor{MetricBlue!38}0.408 & \cellcolor{MetricBlue!33}0.346 & \cellcolor{MetricBlue!20}0.190 & \cellcolor{MetricBlue!58}0.659 & \cellcolor{MetricBlue!61}0.702 & \cellcolor{MetricBlue!51}0.420 \\
        Uberon 2025 & Qwen2.5-14B & Base, existence loop & \cellcolor{MetricBlue!37}0.401 & \cellcolor{MetricBlue!32}0.335 & \cellcolor{MetricBlue!19}0.178 & \cellcolor{MetricBlue!57}0.645 & \cellcolor{MetricBlue!60}0.685 & \cellcolor{MetricBlue!65}0.027 \\
        Uberon 2025 & Qwen2.5-14B & LoRA, no loop & \cellcolor{MetricBlue!39}0.420 & \cellcolor{MetricBlue!34}0.366 & \cellcolor{MetricBlue!21}0.201 & \cellcolor{MetricBlue!59}0.670 & \cellcolor{MetricBlue!61}0.703 & \cellcolor{MetricBlue!65}0.080 \\
        Uberon 2025 & Qwen2.5-14B & LoRA, existence loop & \cellcolor{MetricBlue!39}0.421 & \cellcolor{MetricBlue!34}0.367 & \cellcolor{MetricBlue!21}0.202 & \cellcolor{MetricBlue!59}0.670 & \cellcolor{MetricBlue!61}0.703 & \cellcolor{MetricBlue!65}0.009 \\
        Uberon 2025 & Llama-3.1-8B & Base, no loop & \cellcolor{MetricBlue!36}0.393 & \cellcolor{MetricBlue!31}0.319 & \cellcolor{MetricBlue!19}0.173 & \cellcolor{MetricBlue!57}0.653 & \cellcolor{MetricBlue!60}0.691 & \cellcolor{MetricBlue!59}0.330 \\
        Uberon 2025 & Llama-3.1-8B & Base, existence loop & \cellcolor{MetricBlue!37}0.404 & \cellcolor{MetricBlue!31}0.322 & \cellcolor{MetricBlue!19}0.178 & \cellcolor{MetricBlue!57}0.654 & \cellcolor{MetricBlue!60}0.693 & \cellcolor{MetricBlue!65}0.036 \\
        Uberon 2025 & Llama-3.1-8B & LoRA, no loop & \cellcolor{MetricBlue!40}0.436 & \cellcolor{MetricBlue!33}0.344 & \cellcolor{MetricBlue!20}0.192 & \cellcolor{MetricBlue!59}0.678 & \cellcolor{MetricBlue!62}0.712 & \cellcolor{MetricBlue!61}0.304 \\
        Uberon 2025 & Llama-3.1-8B & LoRA, existence loop & \cellcolor{MetricBlue!40}0.435 & \cellcolor{MetricBlue!33}0.346 & \cellcolor{MetricBlue!21}0.195 & \cellcolor{MetricBlue!59}0.675 & \cellcolor{MetricBlue!62}0.709 & \cellcolor{MetricBlue!65}0.089 \\
        \midrule
        Uberon-Constraint-Stress & Qwen2.5-14B & Base, no loop & \cellcolor{MetricBlue!30}0.307 & \cellcolor{MetricBlue!25}0.246 & \cellcolor{MetricBlue!14}0.115 & \cellcolor{MetricBlue!53}0.602 & \cellcolor{MetricBlue!57}0.644 & \cellcolor{MetricBlue!43}0.523 \\
        Uberon-Constraint-Stress & Qwen2.5-14B & Base, existence loop & \cellcolor{MetricBlue!30}0.307 & \cellcolor{MetricBlue!25}0.247 & \cellcolor{MetricBlue!14}0.116 & \cellcolor{MetricBlue!53}0.602 & \cellcolor{MetricBlue!57}0.644 & \cellcolor{MetricBlue!65}0.025 \\
        Uberon-Constraint-Stress & Qwen2.5-14B & LoRA, no loop & \cellcolor{MetricBlue!38}0.416 & \cellcolor{MetricBlue!35}0.379 & \cellcolor{MetricBlue!24}0.242 & \cellcolor{MetricBlue!59}0.674 & \cellcolor{MetricBlue!62}0.712 & \cellcolor{MetricBlue!53}0.403 \\
        Uberon-Constraint-Stress & Qwen2.5-14B & LoRA, existence loop & \cellcolor{MetricBlue!38}0.410 & \cellcolor{MetricBlue!35}0.373 & \cellcolor{MetricBlue!24}0.237 & \cellcolor{MetricBlue!59}0.669 & \cellcolor{MetricBlue!62}0.707 & \cellcolor{MetricBlue!65}0.024 \\
        Uberon-Constraint-Stress & Llama-3.1-8B & Base, no loop & \cellcolor{MetricBlue!31}0.320 & \cellcolor{MetricBlue!26}0.257 & \cellcolor{MetricBlue!14}0.117 & \cellcolor{MetricBlue!53}0.603 & \cellcolor{MetricBlue!57}0.644 & \cellcolor{MetricBlue!60}0.310 \\
        Uberon-Constraint-Stress & Llama-3.1-8B & Base, existence loop & \cellcolor{MetricBlue!31}0.319 & \cellcolor{MetricBlue!25}0.255 & \cellcolor{MetricBlue!14}0.116 & \cellcolor{MetricBlue!53}0.603 & \cellcolor{MetricBlue!57}0.644 & \cellcolor{MetricBlue!65}0.023 \\
        Uberon-Constraint-Stress & Llama-3.1-8B & LoRA, no loop & \cellcolor{MetricBlue!38}0.409 & \cellcolor{MetricBlue!34}0.368 & \cellcolor{MetricBlue!23}0.226 & \cellcolor{MetricBlue!58}0.664 & \cellcolor{MetricBlue!61}0.702 & \cellcolor{MetricBlue!62}0.285 \\
        Uberon-Constraint-Stress & Llama-3.1-8B & LoRA, existence loop & \cellcolor{MetricBlue!37}0.404 & \cellcolor{MetricBlue!34}0.363 & \cellcolor{MetricBlue!23}0.220 & \cellcolor{MetricBlue!58}0.657 & \cellcolor{MetricBlue!61}0.695 & \cellcolor{MetricBlue!65}0.061 \\
        \bottomrule
    \end{tabular}}
\end{table*}

\begin{figure*}[t]
    \centering
    \includegraphics[width=\linewidth]{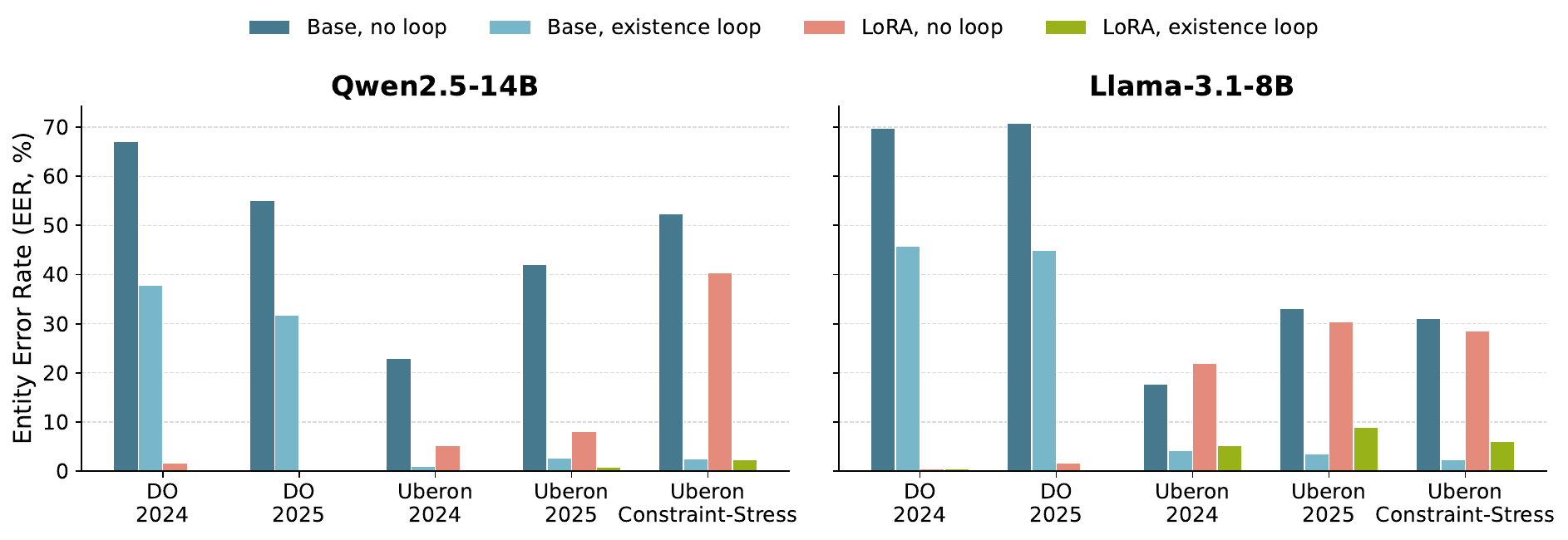}
    \caption{Entity Error Rate (EER) before and after entity-existence repair.
    Lower is better.
    The repair loop substantially reduces unresolved-entity failures for base
    models and for difficult Uberon settings.}
    \label{fig:result:eer}
\end{figure*}

\Cref{fig:result:eer} highlights the EER reductions across the five test
settings.  Base models often produce many unresolved ontology entities,
especially on DO.  For Qwen2.5-14B, the loop reduces EER from 67.0\% to
37.8\% on DO 2024, from 55.1\% to 31.8\% on DO 2025, and from 52.3\% to
2.5\% on Uberon-Constraint-Stress.  Llama-3.1-8B shows similar reductions,
including 69.7\% to 45.7\% on DO 2024 and 31.0\% to 2.3\% on
Uberon-Constraint-Stress.  These results support the central motivation of the
workflow: direct generation often produces entity-grounding failures, and a
targeted QC loop can remove many invalid outputs before curator review.

LoRA fine-tuning generally improves conventional text metrics, but it does
not always remove entity errors.  The clearest improvement appears on DO,
where fine-tuned models achieve higher lexical and semantic similarity while
keeping EER near zero.  On Uberon, fine-tuning improves textual similarity but
can still leave substantial knowledge-compatibility failures.  For example,
Llama-3.1-8B with LoRA still has EER values of 5.2\%, 8.9\%, and 6.1\% after
entity-existence repair on
Uberon 2024, Uberon 2025, and Uberon-Constraint-Stress,
respectively.  This result motivates separate QC checks beyond text metrics.

For completeness, \Cref{tab:app:all_loop_text} reports the corresponding
conventional text-generation scores when direct generation is compared with
the full all-loop workflow.

\subsection{Logical Inconsistency Results}\label{sec:results:logic}

\begin{table*}[t]
    \centering
    \caption{Logical Inconsistency Rate (LIR) after entity-existence
    repair.  The comparison is between existence-loop and all-loop
    settings; all-loop adds the logical repair loop on top of the
    existence-loop setting, so both share the same entity-grounded
    candidate pool.
    Real count reports the number of inconsistent candidates over the
    number of entity-grounded candidates.  Bold all-loop values indicate
    a lower LIR than the paired existence-loop setting.}
    \label{tab:result:lir}
    \setlength{\tabcolsep}{4pt}
    \begin{tabular}{lllcc}
        \toprule
        Test setting & Model & Loop setting & LIR \(\downarrow\) & Real count \\
        \midrule
        Uberon 2024 & Qwen2.5-14B & Base, existence loop & 1.05\% & 1/95 \\
        Uberon 2024 & Qwen2.5-14B & Base, all loops & 1.05\% & 1/95 \\
        Uberon 2024 & Qwen2.5-14B & LoRA, existence loop & 2.08\% & 2/96 \\
        Uberon 2024 & Qwen2.5-14B & LoRA, all loops & 2.08\% & 2/96 \\
        Uberon 2024 & Llama-3.1-8B & Base, existence loop & 0.00\% & 0/92 \\
        Uberon 2024 & Llama-3.1-8B & Base, all loops & 0.00\% & 0/92 \\
        Uberon 2024 & Llama-3.1-8B & LoRA, existence loop & 0.00\% & 0/91 \\
        Uberon 2024 & Llama-3.1-8B & LoRA, all loops & 0.00\% & 0/91 \\
        \midrule
        Uberon 2025 & Qwen2.5-14B & Base, existence loop & 0.92\% & 1/109 \\
        Uberon 2025 & Qwen2.5-14B & Base, all loops & 0.92\% & 1/109 \\
        Uberon 2025 & Qwen2.5-14B & LoRA, existence loop & 1.80\% & 2/111 \\
        Uberon 2025 & Qwen2.5-14B & LoRA, all loops & \textbf{0.00\%} & \textbf{0/111} \\
        Uberon 2025 & Llama-3.1-8B & Base, existence loop & 0.93\% & 1/108 \\
        Uberon 2025 & Llama-3.1-8B & Base, all loops & \textbf{0.00\%} & \textbf{0/108} \\
        Uberon 2025 & Llama-3.1-8B & LoRA, existence loop & 0.98\% & 1/102 \\
        Uberon 2025 & Llama-3.1-8B & LoRA, all loops & 0.98\% & 1/102 \\
        \midrule
        Uberon-Constraint-Stress & Qwen2.5-14B & Base, existence loop & 0.42\% & 4/949 \\
        Uberon-Constraint-Stress & Qwen2.5-14B & Base, all loops & \textbf{0.32\%} & \textbf{3/949} \\
        Uberon-Constraint-Stress & Qwen2.5-14B & LoRA, existence loop & 2.95\% & 28/950 \\
        Uberon-Constraint-Stress & Qwen2.5-14B & LoRA, all loops & \textbf{0.42\%} & \textbf{4/950} \\
        Uberon-Constraint-Stress & Llama-3.1-8B & Base, existence loop & 1.26\% & 12/951 \\
        Uberon-Constraint-Stress & Llama-3.1-8B & Base, all loops & \textbf{0.21\%} & \textbf{2/951} \\
        Uberon-Constraint-Stress & Llama-3.1-8B & LoRA, existence loop & 1.20\% & 11/914 \\
        Uberon-Constraint-Stress & Llama-3.1-8B & LoRA, all loops & \textbf{1.09\%} & \textbf{10/914} \\
        \bottomrule
    \end{tabular}
\end{table*}

Logical reasoning is applied after entity existence has been checked.  This
ordering is natural for ontology QC: generated axioms should first be grounded
to existing predicates and object entities before testing whether they are
logically compatible with the ontology.  Therefore, the LIR experiments are
performed on candidates that have passed the entity-existence check after the
EER repair loop.

\Cref{tab:result:lir} summarizes LIR after the entity-existence loop has been
applied for the Uberon settings.  We compare the existence-loop setting with
the all-loop setting, where all loops means that the logical repair loop is
added after the existence loop.  These two settings share the same
entity-grounded candidate pool, whereas no-loop runs
can have a different LIR denominator.  DO shows no reasoner-detected logical
inconsistencies in these runs, consistent with the earlier observation that
the 2023 DO release contains very few terms involved in strong constraint
axioms.  We therefore omit the all-zero DO rows from the table.  Uberon, in
contrast, has non-zero LIR values in both incremental and stress settings.
Adding the logic loop reduces LIR in
some settings, such as Qwen2.5-14B with LoRA on Uberon 2025 and
Uberon-Constraint-Stress, but the reductions are smaller and less consistent
than the EER reductions.  This shows why LIR is needed: text metrics and
entity-resolution checks do not tell us whether generated axioms are logically
compatible with the ontology.  Because different test settings and models
produce different numbers of candidates that pass entity-existence checking,
the LIR denominator also differs across rows.  We therefore report LIR in a
table rather than using a grouped bar chart.

\subsection{Definition Generator Model Selection}
\label{sec:results:model-selection}

Before running the full QC workflow, we evaluated candidate definition
generators on the DO 2024 and DO 2025 test sets using
conventional text-generation metrics.  This preliminary experiment was used
to select the model families for the main experiments.  We first screened
candidate models under direct inference and then fine-tuned stronger
direct-inference candidates for further evaluation.  The model-selection and
input-field ablation experiments use greedy decoding to make the comparisons
deterministic; the decoding setting is reported in
\Cref{tab:app:hyperparameters}.  They are also evaluated on definition-only
DO test sets, which contain all terms with reference definitions:
326 records for DO 2024 and 305 records for DO 2025.  These sets are larger
than the main workflow test sets because the full QC workflow additionally
requires relationship axioms.
\Cref{tab:result:do_model_selection}
reports both direct base-model inference and fine-tuned inference.  Model
names prefixed with FT denote fine-tuned models; other rows use direct
inference from the base model.  We include the knowledge-cutoff information as
a reference for possible data leakage.

\begin{table*}[t]
    \centering
    \scriptsize
    \caption{Definition-generation model selection on DO release-based
    test sets.  FT-prefixed models are evaluated after fine-tuning;
    other models are evaluated by direct base-model inference.  The
    knowledge-cutoff column is included as a data-leakage reference;
    bold marks the best score within each test set.}
    \label{tab:result:do_model_selection}
    \setlength{\tabcolsep}{2.5pt}
    \resizebox{\linewidth}{!}{%
    \begin{tabular}{lllccccc}
        \toprule
        Test set & Model & Knowledge cutoff & Token F1 & ROUGE-L & BLEU-2 & BERT-F1 & Cosine \\
        \midrule
        DO 2024 & Qwen2.5-7B-Instruct & not reported; card Jan. 2025 & 0.280 & 0.184 & 0.084 & 0.587 & 0.705 \\
        DO 2024 & Llama-3.1-8B & 2023.12 & 0.325 & 0.238 & 0.124 & 0.603 & 0.663 \\
        DO 2024 & Llama-3.1-8B-Instruct & 2023.12 & 0.297 & 0.180 & 0.085 & 0.565 & 0.696 \\
        DO 2024 & Qwen2.5-14B-Instruct & not reported; card Jan. 2025 & 0.281 & 0.192 & 0.089 & 0.597 & 0.704 \\
        DO 2024 & gpt-oss-20b & not reported; card Aug. 2025 & 0.239 & 0.144 & 0.063 & 0.525 & 0.536 \\
        DO 2024 & DeepSeek-R1-Distill-Llama-70B & not reported; card Jan. 2025 & 0.267 & 0.163 & 0.077 & 0.569 & 0.626 \\
        DO 2024 & FT-Llama-3.1-8B-Instruct & 2023.12 & \textbf{0.606} & 0.585 & \textbf{0.390} & 0.807 & 0.782 \\
        DO 2024 & FT-Qwen2.5-14B-Instruct & not reported; card Jan. 2025 & 0.602 & \textbf{0.586} & 0.386 & \textbf{0.808} & \textbf{0.786} \\
        DO 2024 & FT-gpt-oss-20b & not reported; card Aug. 2025 & 0.499 & 0.470 & 0.290 & 0.749 & 0.761 \\
        \midrule
        DO 2025 & Qwen2.5-7B-Instruct & not reported; card Jan. 2025 & 0.279 & 0.181 & 0.085 & 0.579 & 0.676 \\
        DO 2025 & Llama-3.1-8B & 2023.12 & 0.332 & 0.246 & 0.132 & 0.600 & 0.670 \\
        DO 2025 & Llama-3.1-8B-Instruct & 2023.12 & 0.291 & 0.180 & 0.087 & 0.557 & 0.673 \\
        DO 2025 & Qwen2.5-14B-Instruct & not reported; card Jan. 2025 & 0.277 & 0.189 & 0.090 & 0.587 & 0.670 \\
        DO 2025 & gpt-oss-20b & not reported; card Aug. 2025 & 0.234 & 0.141 & 0.065 & 0.516 & 0.533 \\
        DO 2025 & DeepSeek-R1-Distill-Llama-70B & not reported; card Jan. 2025 & 0.266 & 0.166 & 0.081 & 0.562 & 0.607 \\
        DO 2025 & FT-Llama-3.1-8B-Instruct & 2023.12 & \textbf{0.599} & 0.588 & \textbf{0.378} & 0.805 & 0.755 \\
        DO 2025 & FT-Qwen2.5-14B-Instruct & not reported; card Jan. 2025 & 0.597 & \textbf{0.589} & 0.373 & \textbf{0.808} & \textbf{0.758} \\
        DO 2025 & FT-gpt-oss-20b & not reported; card Aug. 2025 & 0.515 & 0.501 & 0.310 & 0.760 & 0.730 \\
        \bottomrule
    \end{tabular}}
\end{table*}

The model-selection results show a large gap between direct base-model
inference and fine-tuned inference.  On both DO test sets, the fine-tuned
Qwen2.5-14B and Llama-3.1-8B models obtain the strongest lexical and semantic
scores, whereas the other fine-tuned candidate, gpt-oss-20b, remains lower.
We therefore use Qwen2.5-14B and Llama-3.1-8B as the two model families for
the main workflow experiments.

\subsection{Definition Input-Field Ablation}
\label{sec:results:input-ablation}

We also test which ontology fields should be included in the definition
generator input.  This ablation uses the fine-tuned Qwen2.5-14B-Instruct
model on the same definition-only DO 2024 and DO 2025 test sets used for
model selection.
\Cref{tab:result:do_input_ablation} compares full-field input with variants
that remove selected fields.  Full fields include the term id, name, synonyms,
parents, and subsets.

\begin{table*}[t]
    \centering
    \scriptsize
    \caption{Input-field ablation for DO definition generation using
    the fine-tuned Qwen2.5-14B-Instruct model.  Full fields include
    id, name, synonyms, parents, and subsets; each ablation removes
    the listed field or fields from the input.  Bold marks the best
    score within each test set.}
    \label{tab:result:do_input_ablation}
    \setlength{\tabcolsep}{2.6pt}
    \resizebox{\linewidth}{!}{%
    \begin{tabular}{lllccccc}
        \toprule
        Test set & Input setting & Input fields & Token F1 & ROUGE-L & BLEU-2 & BERT-F1 & Cosine \\
        \midrule
        DO 2024 & Full fields & id, name, synonyms, parents, subsets & 0.602 & 0.586 & 0.386 & 0.808 & 0.786 \\
        DO 2024 & No ID & name, synonyms, parents, subsets & 0.601 & 0.585 & 0.381 & 0.807 & \textbf{0.791} \\
        DO 2024 & No ID/synonyms & name, parents, subsets & 0.596 & 0.579 & 0.379 & 0.803 & 0.776 \\
        DO 2024 & No ID/parents & name, synonyms, subsets & 0.576 & 0.555 & 0.348 & 0.795 & 0.756 \\
        DO 2024 & No ID/subsets & name, synonyms, parents & \textbf{0.613} & \textbf{0.597} & \textbf{0.396} & \textbf{0.813} & 0.790 \\
        \midrule
        DO 2025 & Full fields & id, name, synonyms, parents, subsets & 0.597 & 0.589 & 0.373 & 0.808 & 0.758 \\
        DO 2025 & No ID & name, synonyms, parents, subsets & 0.601 & 0.593 & 0.382 & 0.808 & 0.762 \\
        DO 2025 & No ID/synonyms & name, parents, subsets & 0.594 & 0.588 & 0.376 & 0.806 & 0.755 \\
        DO 2025 & No ID/parents & name, synonyms, subsets & 0.575 & 0.568 & 0.346 & 0.799 & 0.731 \\
        DO 2025 & No ID/subsets & name, synonyms, parents & \textbf{0.607} & \textbf{0.600} & \textbf{0.391} & \textbf{0.812} & \textbf{0.769} \\
        \bottomrule
    \end{tabular}}
\end{table*}

The ablation shows that parent context is the most important auxiliary field.
Removing parents produces the largest drop on both DO test sets, reducing
Token F1 from 0.602 to 0.576 in 2024 and from 0.597 to 0.575 in 2025.
Removing synonyms changes performance only slightly, whereas removing subsets
often improves performance; the no-id/subsets setting gives the best Token
F1, ROUGE-L, BLEU-2, and BERT-F1 scores in both years.  Based on these
results, the main workflow uses the class name together with parent names and
parent definitions as the core input fields.  To check whether this input
design is practical, we further inspected the official DO and Uberon release
files used in this study.  Terms without textual definitions or relationship
axioms still retain asserted parent classes in these releases, so parent
context is available even for the target records that require definition
generation.

\subsection{Axiom Extractor Performance}
\label{sec:results:axiom-extractor}

We next evaluate the axiom extractor itself on the no-loop workflow outputs.
This isolates extraction quality before entity-existence and logic repair are
applied.  \Cref{tab:result:axiom_extractor_test} compares five-shot base
prompting with the LoRA fine-tuned axiom extractor on the DO, Uberon, and
Uberon-Constraint-Stress test sets.  All predicate and object strings are
normalized by lowercasing,
replacing underscores with spaces, and collapsing whitespace before
comparison.

The metrics measure different levels of extraction quality.  Parse rate is
the proportion of outputs that can be parsed into axiom records.  Exact match
requires the whole predicted axiom set for a term to match the reference set.
Pair precision, recall, and F1 compare complete predicate--object pairs.
Predicate accuracy and object accuracy evaluate each component separately.
Hit rate is a term-level metric: it counts whether at least one predicted
axiom matches any reference axiom.  The validation-split performance used to
check whether the LoRA extractor learned the axiom-extraction task is reported
in \Cref{tab:app:axiom_extractor_validation}.

\begin{table*}[t]
    \centering
    \scriptsize
    \caption{Axiom-extractor performance on test-set no-loop workflow
    outputs.  Base rows use five-shot prompting, whereas LoRA rows use
    the fine-tuned axiom extractor.  Predicate and object strings are
    normalized before comparison; bold marks the higher score within
    each Base/LoRA pair.}
    \label{tab:result:axiom_extractor_test}
    \setlength{\tabcolsep}{2.2pt}
    \resizebox{\linewidth}{!}{%
    \begin{tabular}{llllccccccccc}
        \toprule
        Ontology & Setting & Model & Extractor & $n$ & Parse & Exact & Pair P & Pair R & Pair F1 & Pred. Acc. & Obj. Acc. & Hit \\
        \midrule
        DO & 2024 & Qwen2.5-14B & Base + few-shot & 188 & 1.000 & 0.074 & 0.103 & 0.164 & 0.127 & 0.137 & 0.350 & 0.170 \\
        DO & 2024 & Qwen2.5-14B & LoRA & 188 & 1.000 & \textbf{0.777} & \textbf{0.630} & \textbf{0.831} & \textbf{0.717} & \textbf{0.628} & \textbf{0.604} & \textbf{0.894} \\
        DO & 2024 & Llama-3.1-8B & Base + few-shot & 188 & 1.000 & 0.005 & 0.021 & 0.075 & 0.033 & 0.044 & 0.156 & 0.080 \\
        DO & 2024 & Llama-3.1-8B & LoRA & 188 & 1.000 & \textbf{0.777} & \textbf{0.679} & \textbf{0.845} & \textbf{0.753} & \textbf{0.686} & \textbf{0.664} & \textbf{0.894} \\
        \midrule
        DO & 2025 & Qwen2.5-14B & Base + few-shot & 236 & 1.000 & 0.199 & 0.186 & 0.305 & 0.231 & 0.202 & 0.347 & 0.369 \\
        DO & 2025 & Qwen2.5-14B & LoRA & 236 & 1.000 & \textbf{0.669} & \textbf{0.691} & \textbf{0.761} & \textbf{0.725} & \textbf{0.637} & \textbf{0.620} & \textbf{0.877} \\
        DO & 2025 & Llama-3.1-8B & Base + few-shot & 236 & 1.000 & 0.030 & 0.065 & 0.182 & 0.095 & 0.072 & 0.208 & 0.220 \\
        DO & 2025 & Llama-3.1-8B & LoRA & 236 & 1.000 & \textbf{0.682} & \textbf{0.712} & \textbf{0.779} & \textbf{0.744} & \textbf{0.648} & \textbf{0.640} & \textbf{0.903} \\
        \midrule
        Uberon & 2024 & Qwen2.5-14B & Base + few-shot & 96 & 1.000 & 0.010 & 0.132 & 0.090 & 0.107 & 0.320 & 0.118 & 0.333 \\
        Uberon & 2024 & Qwen2.5-14B & LoRA & 96 & 1.000 & \textbf{0.031} & \textbf{0.455} & \textbf{0.410} & \textbf{0.431} & \textbf{0.567} & \textbf{0.405} & \textbf{0.823} \\
        Uberon & 2024 & Llama-3.1-8B & Base + few-shot & 96 & 1.000 & 0.000 & 0.096 & 0.087 & 0.091 & 0.323 & 0.098 & 0.312 \\
        Uberon & 2024 & Llama-3.1-8B & LoRA & 96 & 1.000 & \textbf{0.021} & \textbf{0.326} & \textbf{0.351} & \textbf{0.338} & \textbf{0.453} & \textbf{0.286} & \textbf{0.781} \\
        \midrule
        Uberon & 2025 & Qwen2.5-14B & Base + few-shot & 112 & 1.000 & 0.036 & 0.175 & 0.158 & 0.166 & 0.334 & 0.214 & 0.446 \\
        Uberon & 2025 & Qwen2.5-14B & LoRA & 112 & 1.000 & \textbf{0.071} & \textbf{0.251} & \textbf{0.278} & \textbf{0.264} & \textbf{0.449} & \textbf{0.305} & \textbf{0.527} \\
        Uberon & 2025 & Llama-3.1-8B & Base + few-shot & 112 & 1.000 & 0.000 & 0.114 & 0.142 & 0.127 & 0.307 & 0.175 & 0.402 \\
        Uberon & 2025 & Llama-3.1-8B & LoRA & 112 & 1.000 & \textbf{0.062} & \textbf{0.155} & \textbf{0.212} & \textbf{0.179} & \textbf{0.370} & \textbf{0.200} & \textbf{0.509} \\
        \midrule
        Uberon & Constraint-Stress & Qwen2.5-14B & Base + few-shot & 973 & 1.000 & 0.011 & 0.113 & 0.063 & 0.081 & 0.225 & 0.084 & 0.222 \\
        Uberon & Constraint-Stress & Qwen2.5-14B & LoRA & 973 & 1.000 & \textbf{0.046} & \textbf{0.210} & \textbf{0.178} & \textbf{0.193} & \textbf{0.298} & \textbf{0.164} & \textbf{0.470} \\
        Uberon & Constraint-Stress & Llama-3.1-8B & Base + few-shot & 973 & 0.975 & 0.003 & 0.075 & 0.057 & 0.065 & 0.216 & 0.070 & 0.201 \\
        Uberon & Constraint-Stress & Llama-3.1-8B & LoRA & 973 & \textbf{1.000} & \textbf{0.046} & \textbf{0.166} & \textbf{0.179} & \textbf{0.172} & \textbf{0.275} & \textbf{0.147} & \textbf{0.485} \\
        \bottomrule
    \end{tabular}}
\end{table*}

The test-set results show that LoRA fine-tuning substantially improves axiom
extraction over five-shot prompting.  The improvement is especially large for
DO, where LoRA raises hit rate from 0.170 to 0.894 for Qwen2.5-14B on
2024 and from 0.220 to 0.903 for Llama-3.1-8B on 2025.
Fine-tuned LLM extractors achieve high hit rates on most test settings, with
the best setting exceeding 0.90, suggesting that LLM-based axiom extraction
has substantial potential.
Uberon remains harder, especially the constraint-stress subset, but LoRA still
improves pair-level F1 and hit rate for both model families.  These results
justify using a fine-tuned axiom extractor in the workflow while keeping
five-shot base extraction as the no-finetuning comparison.

\section{Discussion}\label{sec:discussion}

The main finding of this study is that LLM-assisted ontology definition
generation should not be evaluated only as a text-generation task.  The
definition text may be fluent and similar to a reference definition, while the
relations implied by that text may still fail entity grounding or violate
ontology-level constraints.  This distinction is visible throughout the
experiments.  Conventional metrics remain useful for measuring lexical and
semantic similarity, especially when comparing base and fine-tuned definition
generators, but they do not expose whether the generated content can be
accepted by an ontology engineering workflow.  EER and LIR therefore provide a
complementary view of generation quality: they measure whether generated
definitions can be connected to existing ontology entities and whether the
resulting structured claims are logically compatible with the ontology.

The results also show that different quality-control modules address
different kinds of failures.  The entity-existence loop produces clear EER
reductions in many settings, particularly for base models and for the
constraint-focused Uberon subset.  This is expected because unresolved
predicates or object entities provide direct local feedback to the next
generation attempt.  The model can be told which relation or object could not
be grounded, and a new candidate can often avoid that specific error.  In
contrast, logical inconsistency is harder to repair.  A candidate may become
unsatisfiable because of global ontology constraints, disjointness axioms,
domain or range restrictions, or interactions among multiple generated
relations.  These failures are sparse and less locally diagnostic, which helps
explain why the logic loop yields only limited reductions in LIR even when EER
repair is effective.

The experiments further suggest that fine-tuning and QC loops play different
roles.  Fine-tuning improves definition-generation quality under conventional
metrics and substantially improves axiom extraction compared with five-shot
prompting.  However, fine-tuning alone does not guarantee ontology
compatibility.  In several Uberon settings, LoRA models retain non-zero EER
after repair, and LIR remains observable even when text scores are strong.
This supports a layered workflow: use fine-tuned models to improve generation
and extraction quality, but keep explicit symbolic checks before any generated
candidate is considered suitable for curator review.

The stress-test results also clarify why the choice of ontology affects what
can be learned from an evaluation.  The DO splits reflect a realistic
incremental maintenance setting, where new or newly defined terms are added
across releases.  However, because only a small number of DO terms are involved
in strong constraint axioms, this setting provides limited evidence about
reasoner-detected logical inconsistency.  Uberon complements DO in this
respect: its regular incremental splits evaluate the same maintenance scenario
in a more axiom-rich ontology, and the Uberon-Constraint-Stress subset
intentionally concentrates terms that are more likely to exercise disjointness
and other strong constraints.  The three datasets therefore serve different
purposes.

Although this paper instantiates EER and LIR for biomedical ontologies, the
underlying idea is broader.  Many structured-knowledge generation tasks
require generated text or claims to align with a controlled vocabulary,
schema, knowledge graph, database, or rule system.  In such settings, the
central question is not only whether the output sounds plausible, but whether
it can be grounded and checked against the target knowledge system.  The
workflow studied here provides one concrete implementation of this principle:
LLMs propose content, while external knowledge-system checks decide whether
the content can be accepted, repaired, or flagged.  Future work should test
the same incompatibility metrics in other structured domains.  More broadly,
similar checks could be applied to general LLM text-generation tasks by
measuring whether generated entities can be grounded in existing knowledge
resources and whether the generated claims remain logically consistent with
established knowledge.

\section{Conclusion}\label{sec:conclusion}

This paper studied LLM-assisted ontology definition generation as an
knowledge-compatibility problem instantiated in biomedical ontologies, rather
than only as a text-generation problem.  The results show that LLMs can
produce useful ontology-definition candidates, but they also demonstrate that
fluent, reference-like definitions are not necessarily compatible with the
target knowledge system.  A generated definition may still contain unresolved
entities or introduce logical inconsistencies that are invisible to
conventional text-similarity metrics.

To address this gap, we introduced EER and LIR as structured-knowledge
incompatibility metrics for evaluating entity grounding and logical consistency,
and developed an agentic QC framework that combines generation, grounding
verification, reasoning, and iterative repair.  Experiments on DO, Uberon, and
Uberon-Constraint-Stress show that the workflow substantially reduces
entity-grounding failures while preserving strong conventional
text-generation performance.  More broadly, this work suggests that LLM
generation in knowledge-intensive domains should be evaluated not only by
linguistic quality, but also by whether generated content can be grounded,
checked, and integrated into the target knowledge system.

\section*{Acknowledgments}
This work was supported by the Department of Education of Guangdong Province
(2025ZDZX2065) and the Shenzhen Science and Technology Program
(JCYJ202412\allowbreak021302\allowbreak12016).  The funders had no role in
study design, data
collection and analysis, decision to publish, or preparation of the manuscript.

\section*{Data and Code Availability}
The source code and constructed evaluation datasets supporting this study are
publicly available at
\url{https://github.com/iwcyou/DO-DefinitionGeneration}.

\section*{Declaration of Competing Interest}
The authors declare that they have no known competing financial interests or
personal relationships that could have appeared to influence the work reported
in this paper.

\FloatBarrier
\bibliographystyle{cas-model2-names}
\bibliography{references}

\clearpage
\appendix
\hypersetup{pageanchor=false}
\setcounter{page}{1}
\setcounter{figure}{0}
\renewcommand{\thefigure}{\Alph{section}.\arabic{figure}}
\renewcommand{\theHfigure}{appendix.\Alph{section}.\arabic{figure}}
\setcounter{table}{0}
\renewcommand{\thetable}{\Alph{section}.\arabic{table}}
\renewcommand{\theHtable}{appendix.\Alph{section}.\arabic{table}}
\section{%
    Additional Experimental Results
}\label{app:results}

\subsection{Fine-Tuning and Decoding Hyperparameters}
\label{app:results:hyperparameters}

\Cref{tab:app:hyperparameters} summarizes the LoRA and supervised
fine-tuning (SFT) settings, the greedy decoding setting used for
definition-generator model selection and input-field ablation, and the
random-sampling parameters used in the
repair-loop experiments.  The same LoRA configuration is used for definition
generation and axiom extraction.  Greedy decoding is used for the
definition-generation comparison experiments to make model and field
comparisons deterministic.  In the random-sampling repair experiments, the
sampling parameters are applied to regenerated definitions and corrected
axioms inside the repair loops.

\begin{table*}[t]
    \centering
    \small
    \caption{Fine-tuning, greedy definition-generation comparison, and
    random-sampling repair settings used in the experiments.  The Low-Rank
    Adaptation (LoRA) and supervised fine-tuning (SFT) settings are shared by
    the definition generator and axiom extractor unless otherwise noted.}
    \label{tab:app:hyperparameters}
    \setlength{\tabcolsep}{4pt}
    \begin{tabular}{lll}
        \toprule
        Setting & Parameter & Value \\
        \midrule
        LoRA/SFT & Base models & Qwen2.5-14B-Instruct; Llama-3.1-8B-Instruct \\
        LoRA/SFT & LoRA rank \(r\) & 64 \\
        LoRA/SFT & LoRA alpha & 128 \\
        LoRA/SFT & LoRA dropout & 0.05 \\
        LoRA/SFT & Target modules & q, k, v, o, gate, up, down projections \\
        LoRA/SFT & Rank-stabilized LoRA (RS-LoRA) & enabled \\
        LoRA/SFT & Training epochs & 5 \\
        LoRA/SFT & Train / evaluation batch size & 4 / 4 per device \\
        LoRA/SFT & Gradient accumulation & 4 steps \\
        LoRA/SFT & Learning rate & \(1\times10^{-4}\) \\
        LoRA/SFT & Scheduler / warmup & cosine / 0.05 \\
        LoRA/SFT & Optimizer & \texttt{adamw\_torch\_fused} \\
        LoRA/SFT & Weight decay / max grad norm & 0.01 / 1.0 \\
        LoRA/SFT & Precision & bfloat16 (BF16) training and evaluation \\
        LoRA/SFT & Max sequence length & 2048 tokens \\
        LoRA/SFT & Save / evaluation / logging steps & 100 / 100 / 10 \\
        LoRA/SFT & Random seed & 42 \\
        \midrule
        Greedy definition-generation comparison & \texttt{do\_sample} & \texttt{False} \\
        Greedy definition-generation comparison & Temperature & not used \\
        Greedy definition-generation comparison & top-\(p\) / top-\(k\) & not used / not used \\
        Greedy definition-generation comparison & Max new tokens & 128 \\
        \midrule
        Random-sampling repair & \texttt{do\_sample} & \texttt{True} \\
        Random-sampling repair & Temperature & 0.6 \\
        Random-sampling repair & top-\(p\) & 0.9 \\
        Random-sampling repair & top-\(k\) & 40 \\
        Random-sampling repair & Existence-loop budget & 5 retries per failed record \\
        Random-sampling repair & Logic-loop budget & 5 axiom-correction attempts \\
        Random-sampling repair & Few-shot axiom examples & 5 \\
        \bottomrule
    \end{tabular}
\end{table*}

\subsection{Full-Workflow Text-Generation Scores}
\label{app:results:text-generation}

\Cref{tab:app:all_loop_text} compares direct generation with the full
all-loop workflow using both conventional text-generation metrics and EER.
The all-loop workflow usually preserves the lexical and semantic scores of
direct generation, while reducing EER in many settings.  This supports the
interpretation that the QC loops mainly filter or repair
knowledge-compatibility failures rather than trading away conventional text
quality.

\begin{table*}[t]
    \centering
    \scriptsize
    \caption{Conventional text-generation scores and Entity Error Rate
    (EER) comparing direct generation with the full QC workflow.
    Darker cells indicate better values: higher conventional scores
    and lower EER.  Arrows in column headers indicate the preferred
    metric direction.}
    \label{tab:app:all_loop_text}
    \setlength{\tabcolsep}{3pt}
    \resizebox{\linewidth}{!}{%
    \begin{tabular}{lllcccccc}
        \toprule
        Test setting & Model & Loop setting & Token F1 \(\uparrow\) & ROUGE-L \(\uparrow\) & BLEU-2 \(\uparrow\) & BERT-F1 \(\uparrow\) & Cosine \(\uparrow\) & EER \(\downarrow\) \\
        \midrule
        DO 2024 & Qwen2.5-14B & Base, no loop & \cellcolor{MetricBlue!28}0.283 & \cellcolor{MetricBlue!23}0.222 & \cellcolor{MetricBlue!14}0.107 & \cellcolor{MetricBlue!54}0.608 & \cellcolor{MetricBlue!58}0.664 & \cellcolor{MetricBlue!31}0.670 \\
        DO 2024 & Qwen2.5-14B & Base, all loops & \cellcolor{MetricBlue!28}0.293 & \cellcolor{MetricBlue!24}0.236 & \cellcolor{MetricBlue!14}0.118 & \cellcolor{MetricBlue!54}0.619 & \cellcolor{MetricBlue!59}0.676 & \cellcolor{MetricBlue!55}0.378 \\
        DO 2024 & Qwen2.5-14B & LoRA, no loop & \cellcolor{MetricBlue!52}0.588 & \cellcolor{MetricBlue!51}0.579 & \cellcolor{MetricBlue!35}0.380 & \cellcolor{MetricBlue!63}0.731 & \cellcolor{MetricBlue!65}0.766 & \cellcolor{MetricBlue!65}0.016 \\
        DO 2024 & Qwen2.5-14B & LoRA, all loops & \cellcolor{MetricBlue!52}0.588 & \cellcolor{MetricBlue!51}0.579 & \cellcolor{MetricBlue!35}0.381 & \cellcolor{MetricBlue!63}0.731 & \cellcolor{MetricBlue!65}0.766 & \cellcolor{MetricBlue!65}0.000 \\
        DO 2024 & Llama-3.1-8B & Base, no loop & \cellcolor{MetricBlue!30}0.310 & \cellcolor{MetricBlue!24}0.240 & \cellcolor{MetricBlue!14}0.115 & \cellcolor{MetricBlue!52}0.593 & \cellcolor{MetricBlue!56}0.642 & \cellcolor{MetricBlue!29}0.697 \\
        DO 2024 & Llama-3.1-8B & Base, all loops & \cellcolor{MetricBlue!31}0.320 & \cellcolor{MetricBlue!25}0.244 & \cellcolor{MetricBlue!15}0.123 & \cellcolor{MetricBlue!54}0.608 & \cellcolor{MetricBlue!58}0.656 & \cellcolor{MetricBlue!48}0.457 \\
        DO 2024 & Llama-3.1-8B & LoRA, no loop & \cellcolor{MetricBlue!51}0.580 & \cellcolor{MetricBlue!50}0.562 & \cellcolor{MetricBlue!35}0.372 & \cellcolor{MetricBlue!63}0.725 & \cellcolor{MetricBlue!65}0.764 & \cellcolor{MetricBlue!65}0.005 \\
        DO 2024 & Llama-3.1-8B & LoRA, all loops & \cellcolor{MetricBlue!51}0.580 & \cellcolor{MetricBlue!50}0.562 & \cellcolor{MetricBlue!35}0.372 & \cellcolor{MetricBlue!63}0.725 & \cellcolor{MetricBlue!65}0.764 & \cellcolor{MetricBlue!65}0.005 \\
        \midrule
        DO 2025 & Qwen2.5-14B & Base, no loop & \cellcolor{MetricBlue!31}0.327 & \cellcolor{MetricBlue!25}0.253 & \cellcolor{MetricBlue!16}0.143 & \cellcolor{MetricBlue!56}0.636 & \cellcolor{MetricBlue!60}0.688 & \cellcolor{MetricBlue!41}0.551 \\
        DO 2025 & Qwen2.5-14B & Base, all loops & \cellcolor{MetricBlue!32}0.331 & \cellcolor{MetricBlue!26}0.260 & \cellcolor{MetricBlue!17}0.148 & \cellcolor{MetricBlue!56}0.636 & \cellcolor{MetricBlue!60}0.687 & \cellcolor{MetricBlue!60}0.318 \\
        DO 2025 & Qwen2.5-14B & LoRA, no loop & \cellcolor{MetricBlue!53}0.606 & \cellcolor{MetricBlue!53}0.602 & \cellcolor{MetricBlue!37}0.399 & \cellcolor{MetricBlue!64}0.739 & \cellcolor{MetricBlue!65}0.771 & \cellcolor{MetricBlue!65}0.000 \\
        DO 2025 & Qwen2.5-14B & LoRA, all loops & \cellcolor{MetricBlue!53}0.606 & \cellcolor{MetricBlue!53}0.602 & \cellcolor{MetricBlue!37}0.399 & \cellcolor{MetricBlue!64}0.739 & \cellcolor{MetricBlue!65}0.771 & \cellcolor{MetricBlue!65}0.000 \\
        DO 2025 & Llama-3.1-8B & Base, no loop & \cellcolor{MetricBlue!33}0.345 & \cellcolor{MetricBlue!26}0.257 & \cellcolor{MetricBlue!16}0.139 & \cellcolor{MetricBlue!54}0.611 & \cellcolor{MetricBlue!58}0.658 & \cellcolor{MetricBlue!28}0.708 \\
        DO 2025 & Llama-3.1-8B & Base, all loops & \cellcolor{MetricBlue!33}0.348 & \cellcolor{MetricBlue!26}0.260 & \cellcolor{MetricBlue!17}0.144 & \cellcolor{MetricBlue!54}0.615 & \cellcolor{MetricBlue!58}0.659 & \cellcolor{MetricBlue!49}0.449 \\
        DO 2025 & Llama-3.1-8B & LoRA, no loop & \cellcolor{MetricBlue!54}0.608 & \cellcolor{MetricBlue!53}0.605 & \cellcolor{MetricBlue!37}0.396 & \cellcolor{MetricBlue!63}0.727 & \cellcolor{MetricBlue!65}0.759 & \cellcolor{MetricBlue!65}0.017 \\
        DO 2025 & Llama-3.1-8B & LoRA, all loops & \cellcolor{MetricBlue!54}0.608 & \cellcolor{MetricBlue!53}0.605 & \cellcolor{MetricBlue!37}0.396 & \cellcolor{MetricBlue!63}0.727 & \cellcolor{MetricBlue!65}0.759 & \cellcolor{MetricBlue!65}0.000 \\
        \midrule
        Uberon 2024 & Qwen2.5-14B & Base, no loop & \cellcolor{MetricBlue!39}0.428 & \cellcolor{MetricBlue!33}0.353 & \cellcolor{MetricBlue!21}0.203 & \cellcolor{MetricBlue!58}0.663 & \cellcolor{MetricBlue!62}0.717 & \cellcolor{MetricBlue!65}0.229 \\
        Uberon 2024 & Qwen2.5-14B & Base, all loops & \cellcolor{MetricBlue!39}0.425 & \cellcolor{MetricBlue!33}0.351 & \cellcolor{MetricBlue!21}0.200 & \cellcolor{MetricBlue!58}0.664 & \cellcolor{MetricBlue!62}0.717 & \cellcolor{MetricBlue!65}0.010 \\
        Uberon 2024 & Qwen2.5-14B & LoRA, no loop & \cellcolor{MetricBlue!39}0.428 & \cellcolor{MetricBlue!35}0.374 & \cellcolor{MetricBlue!23}0.228 & \cellcolor{MetricBlue!61}0.699 & \cellcolor{MetricBlue!64}0.742 & \cellcolor{MetricBlue!65}0.052 \\
        Uberon 2024 & Qwen2.5-14B & LoRA, all loops & \cellcolor{MetricBlue!39}0.431 & \cellcolor{MetricBlue!35}0.378 & \cellcolor{MetricBlue!23}0.230 & \cellcolor{MetricBlue!61}0.699 & \cellcolor{MetricBlue!64}0.742 & \cellcolor{MetricBlue!65}0.000 \\
        Uberon 2024 & Llama-3.1-8B & Base, no loop & \cellcolor{MetricBlue!38}0.409 & \cellcolor{MetricBlue!32}0.332 & \cellcolor{MetricBlue!19}0.175 & \cellcolor{MetricBlue!57}0.649 & \cellcolor{MetricBlue!61}0.698 & \cellcolor{MetricBlue!65}0.177 \\
        Uberon 2024 & Llama-3.1-8B & Base, all loops & \cellcolor{MetricBlue!38}0.409 & \cellcolor{MetricBlue!31}0.331 & \cellcolor{MetricBlue!19}0.176 & \cellcolor{MetricBlue!57}0.650 & \cellcolor{MetricBlue!61}0.698 & \cellcolor{MetricBlue!65}0.042 \\
        Uberon 2024 & Llama-3.1-8B & LoRA, no loop & \cellcolor{MetricBlue!43}0.478 & \cellcolor{MetricBlue!39}0.421 & \cellcolor{MetricBlue!25}0.249 & \cellcolor{MetricBlue!61}0.695 & \cellcolor{MetricBlue!64}0.735 & \cellcolor{MetricBlue!65}0.219 \\
        Uberon 2024 & Llama-3.1-8B & LoRA, all loops & \cellcolor{MetricBlue!44}0.484 & \cellcolor{MetricBlue!39}0.424 & \cellcolor{MetricBlue!26}0.257 & \cellcolor{MetricBlue!61}0.701 & \cellcolor{MetricBlue!64}0.743 & \cellcolor{MetricBlue!65}0.052 \\
        \midrule
        Uberon 2025 & Qwen2.5-14B & Base, no loop & \cellcolor{MetricBlue!38}0.408 & \cellcolor{MetricBlue!33}0.346 & \cellcolor{MetricBlue!20}0.190 & \cellcolor{MetricBlue!58}0.659 & \cellcolor{MetricBlue!61}0.702 & \cellcolor{MetricBlue!51}0.420 \\
        Uberon 2025 & Qwen2.5-14B & Base, all loops & \cellcolor{MetricBlue!37}0.401 & \cellcolor{MetricBlue!32}0.335 & \cellcolor{MetricBlue!19}0.178 & \cellcolor{MetricBlue!57}0.645 & \cellcolor{MetricBlue!60}0.685 & \cellcolor{MetricBlue!65}0.027 \\
        Uberon 2025 & Qwen2.5-14B & LoRA, no loop & \cellcolor{MetricBlue!39}0.420 & \cellcolor{MetricBlue!34}0.366 & \cellcolor{MetricBlue!21}0.201 & \cellcolor{MetricBlue!59}0.670 & \cellcolor{MetricBlue!61}0.703 & \cellcolor{MetricBlue!65}0.080 \\
        Uberon 2025 & Qwen2.5-14B & LoRA, all loops & \cellcolor{MetricBlue!39}0.421 & \cellcolor{MetricBlue!34}0.367 & \cellcolor{MetricBlue!21}0.202 & \cellcolor{MetricBlue!59}0.670 & \cellcolor{MetricBlue!61}0.703 & \cellcolor{MetricBlue!65}0.009 \\
        Uberon 2025 & Llama-3.1-8B & Base, no loop & \cellcolor{MetricBlue!36}0.393 & \cellcolor{MetricBlue!31}0.319 & \cellcolor{MetricBlue!19}0.173 & \cellcolor{MetricBlue!57}0.653 & \cellcolor{MetricBlue!60}0.691 & \cellcolor{MetricBlue!59}0.330 \\
        Uberon 2025 & Llama-3.1-8B & Base, all loops & \cellcolor{MetricBlue!37}0.404 & \cellcolor{MetricBlue!31}0.322 & \cellcolor{MetricBlue!19}0.178 & \cellcolor{MetricBlue!57}0.654 & \cellcolor{MetricBlue!60}0.693 & \cellcolor{MetricBlue!65}0.036 \\
        Uberon 2025 & Llama-3.1-8B & LoRA, no loop & \cellcolor{MetricBlue!40}0.436 & \cellcolor{MetricBlue!33}0.344 & \cellcolor{MetricBlue!20}0.192 & \cellcolor{MetricBlue!59}0.678 & \cellcolor{MetricBlue!62}0.712 & \cellcolor{MetricBlue!61}0.304 \\
        Uberon 2025 & Llama-3.1-8B & LoRA, all loops & \cellcolor{MetricBlue!40}0.435 & \cellcolor{MetricBlue!33}0.346 & \cellcolor{MetricBlue!21}0.195 & \cellcolor{MetricBlue!59}0.675 & \cellcolor{MetricBlue!62}0.709 & \cellcolor{MetricBlue!65}0.089 \\
        \midrule
        Uberon-Constraint-Stress & Qwen2.5-14B & Base, no loop & \cellcolor{MetricBlue!30}0.307 & \cellcolor{MetricBlue!25}0.246 & \cellcolor{MetricBlue!14}0.115 & \cellcolor{MetricBlue!53}0.602 & \cellcolor{MetricBlue!57}0.644 & \cellcolor{MetricBlue!43}0.523 \\
        Uberon-Constraint-Stress & Qwen2.5-14B & Base, all loops & \cellcolor{MetricBlue!30}0.307 & \cellcolor{MetricBlue!25}0.247 & \cellcolor{MetricBlue!14}0.116 & \cellcolor{MetricBlue!53}0.602 & \cellcolor{MetricBlue!57}0.644 & \cellcolor{MetricBlue!65}0.025 \\
        Uberon-Constraint-Stress & Qwen2.5-14B & LoRA, no loop & \cellcolor{MetricBlue!38}0.416 & \cellcolor{MetricBlue!35}0.379 & \cellcolor{MetricBlue!24}0.242 & \cellcolor{MetricBlue!59}0.674 & \cellcolor{MetricBlue!62}0.712 & \cellcolor{MetricBlue!53}0.403 \\
        Uberon-Constraint-Stress & Qwen2.5-14B & LoRA, all loops & \cellcolor{MetricBlue!38}0.410 & \cellcolor{MetricBlue!35}0.373 & \cellcolor{MetricBlue!24}0.237 & \cellcolor{MetricBlue!59}0.669 & \cellcolor{MetricBlue!62}0.707 & \cellcolor{MetricBlue!65}0.024 \\
        Uberon-Constraint-Stress & Llama-3.1-8B & Base, no loop & \cellcolor{MetricBlue!31}0.320 & \cellcolor{MetricBlue!26}0.257 & \cellcolor{MetricBlue!14}0.117 & \cellcolor{MetricBlue!53}0.603 & \cellcolor{MetricBlue!57}0.644 & \cellcolor{MetricBlue!60}0.310 \\
        Uberon-Constraint-Stress & Llama-3.1-8B & Base, all loops & \cellcolor{MetricBlue!31}0.319 & \cellcolor{MetricBlue!25}0.255 & \cellcolor{MetricBlue!14}0.116 & \cellcolor{MetricBlue!53}0.603 & \cellcolor{MetricBlue!57}0.644 & \cellcolor{MetricBlue!65}0.023 \\
        Uberon-Constraint-Stress & Llama-3.1-8B & LoRA, no loop & \cellcolor{MetricBlue!38}0.409 & \cellcolor{MetricBlue!34}0.368 & \cellcolor{MetricBlue!23}0.226 & \cellcolor{MetricBlue!58}0.664 & \cellcolor{MetricBlue!61}0.702 & \cellcolor{MetricBlue!62}0.285 \\
        Uberon-Constraint-Stress & Llama-3.1-8B & LoRA, all loops & \cellcolor{MetricBlue!37}0.404 & \cellcolor{MetricBlue!34}0.363 & \cellcolor{MetricBlue!23}0.220 & \cellcolor{MetricBlue!58}0.657 & \cellcolor{MetricBlue!61}0.695 & \cellcolor{MetricBlue!65}0.061 \\
        \bottomrule
    \end{tabular}}
\end{table*}

\subsection{Axiom Extractor Validation Performance}
\label{app:results:axiom-extractor-validation}

\Cref{tab:app:axiom_extractor_validation} reports validation-split
performance for the LoRA axiom extractor.  This experiment checks whether the
extractor learned to produce parseable predicate--object records before it is
used in the end-to-end workflow.  All validation outputs are parseable, and
the DO validation split shows stronger exact-match and hit-rate performance
than the Uberon validation split.  This difference is consistent with the
test-set results in \Cref{tab:result:axiom_extractor_test}, where Uberon
remains the harder extraction setting because its relationship axioms are more
diverse and constraint-rich.

\begin{table*}[t]
    \centering
    \scriptsize
    \caption{Axiom-extractor performance on validation splits used
    during LoRA training.  The table evaluates whether the fine-tuned
    extractor learned to produce parseable and reference-compatible
    relationship axioms before being used in the workflow.}
    \label{tab:app:axiom_extractor_validation}
    \setlength{\tabcolsep}{2.2pt}
    \resizebox{\linewidth}{!}{%
    \begin{tabular}{llllccccccccc}
        \toprule
        Ontology & Setting & Model & Extractor & $n$ & Parse & Exact & Pair P & Pair R & Pair F1 & Pred. Acc. & Obj. Acc. & Hit \\
        \midrule
        Uberon & Uberon eval & Qwen2.5-14B & LoRA & 1912 & 1.000 & 0.348 & 0.423 & 0.490 & 0.454 & 0.520 & 0.394 & 0.745 \\
        Uberon & Uberon eval & Llama-3.1-8B & LoRA & 1912 & 1.000 & 0.300 & 0.306 & 0.487 & 0.375 & 0.413 & 0.302 & 0.745 \\
        \midrule
        DO & DO eval & Qwen2.5-14B & LoRA & 962 & 1.000 & 0.686 & 0.518 & 0.810 & 0.632 & 0.536 & 0.505 & 0.923 \\
        DO & DO eval & Llama-3.1-8B & LoRA & 962 & 1.000 & 0.682 & 0.606 & 0.797 & 0.689 & 0.635 & 0.588 & 0.915 \\
        \bottomrule
    \end{tabular}}
\end{table*}

\end{document}